\documentclass[sigconf]{acmart}

\copyrightyear{2022}
\acmYear{2022}
\acmConference[WWW '22] {Proceedings of the ACM Web Conference 2022} {April 25--29, 2022}{Lyon, France.}
\acmBooktitle{Proceedings of the ACM Web Conference 2022 (WWW '22), April 25--29, 2022, Lyon, France}
\acmPrice{15.00}
\acmDOI{10.1145/XXXXXX.XXXXXX}

\usepackage{xspace}
\usepackage{enumerate}
\usepackage[ruled]{algorithm2e}
\usepackage{adjustbox}
\usepackage{multirow}
\usepackage{subfigure}
\usepackage{color}
\usepackage{soul}
\usepackage{makecell}
\definecolor{MyRed}{RGB}{151,0,0}
\definecolor{Yellow}{RGB}{243,140,10}

\newcommand{\ourmethod}{\texttt{SUBLIME}\xspace}

\newcommand{\xy}{$\mathbf{X}$, $\mathbf{Y}$}
\newcommand{\xyak}{$\mathbf{X}$, $\mathbf{Y}$, $\mathbf{A}_{knn}$}
\newcommand{\xya}{$\mathbf{X}$, $\mathbf{Y}$, $\mathbf{A}$}

\newcommand{\ak}{$\mathbf{A}_{knn}$}
\newcommand{\onlyx}{$\mathbf{X}$}
\newcommand{\onlya}{$\mathbf{A}$}
\newcommand{\xa}{$\mathbf{X}$, $\mathbf{A}$}

\newcommand{\nameHL}[1]{\underline{\textbf{#1}}}
\newcommand{\fstHL}[1]{\textcolor{MyRed}{\textbf{#1}}}
\newcommand{\secHL}[1]{\textcolor{red}{\underline{#1}}}

\newcommand{\tabincell}[2]{\begin{tabular}{@{}#1@{}}#2\end{tabular}}

\begin{document}
\title{Towards Unsupervised Deep Graph Structure Learning}

\author{Yixin Liu$^{1}$,~~~Yu Zheng$^{2}$,~~~Daokun Zhang$^{1,3}$,~~~Hongxu Chen$^{4}$,~~~Hao Peng$^{5}$,~~~Shirui Pan$^{1}$*}

\makeatletter
\def\authornotetext#1{
	\g@addto@macro\@authornotes{%
	\stepcounter{footnote}\footnotetext{#1}}%
}
\makeatother

\authornotetext{Corresponding author}

\affiliation{%
	\institution{$^1$Monash University \quad $^2$La Trobe University \quad $^3$Monash Suzhou Research Institute \\ $^4$University of Technology Sydney \quad $^5$Beihang University}
	\country{}
}

\email{{yixin.liu, daokun.zhang, shirui.pan}@monash.edu;}
\email{yu.zheng@latrobe.edu.au;   hongxu.chen@uts.edu.au;   penghao@buaa.edu.cn}

\renewcommand{\shortauthors}{Liu et al.}

\begin{abstract}
  In recent years, graph neural networks (GNNs) have emerged as a successful tool in a variety of graph-related applications. However, the performance of GNNs can be deteriorated when noisy connections occur in the original graph structures; besides, the dependence on explicit structures prevents GNNs from being applied to general unstructured scenarios. To address these issues, recently emerged deep graph structure learning (GSL) methods propose to jointly optimize the graph structure along with GNN under the supervision of a node classification task. Nonetheless, these methods focus on a supervised learning scenario, which leads to several problems, i.e., the reliance on labels, the bias of edge distribution, and the limitation on application tasks. In this paper, we propose a more practical GSL paradigm, \textit{unsupervised graph structure learning}, where the learned graph topology is optimized by data itself without any external guidance (i.e., labels). To solve the unsupervised GSL problem, we propose a novel \nameHL{S}tr\nameHL{U}cture \nameHL{B}ootstrapping contrastive \nameHL{L}earn\nameHL{I}ng fra\nameHL{ME}work (\textbf{\ourmethod} for abbreviation) with the aid of self-supervised contrastive learning. Specifically, we generate a learning target from the original data as an ``anchor graph'', and use a contrastive loss to maximize the agreement between the anchor graph and the learned graph. To provide persistent guidance, we design a novel bootstrapping mechanism that upgrades the anchor graph with learned structures during model learning. We also design a series of graph learners and post-processing schemes to model the structures to learn. Extensive experiments on eight benchmark datasets demonstrate the significant effectiveness of our proposed \ourmethod and high quality of the optimized graphs.
\end{abstract}



\maketitle

\section{Introduction}
\label{sec:introduction}
Recent years have witnessed the prosperous development of graph-based applications in numerous domains, such as chemistry, bioinformatics and cybersecurity. As a powerful deep learning tool to model graph-structured data, graph neural networks (GNNs) have drawn increasing attention and achieved state-of-the-art performance in various graph analytical tasks, including node classification \cite{kipf2017semi,velivckovic2018graph}, link prediction \cite{vgae_kipf2016variational,peng2020graph}, and node clustering \cite{daegc_wang2019attributed,agc_zhang2019attributed}. GNNs usually follow a message-passing scheme, where node representations are learned by aggregating information from the neighbors on an observed topology (i.e., the original graph structure). 

Most GNNs rely on a fundamental assumption that the original structure is credible enough to be viewed as ground-truth information for model training. Such assumption, unfortunately, is usually violated in real-world scenarios, since graph structures are usually extracted from complex interaction systems which inevitably contain uncertain, redundant, wrong and missing connections \cite{wang2021graph}. Such noisy information in original topology can seriously damage the performance of GNNs. 
Besides, the reliance on explicit structures hinders GNNs' {broad} applicability. If GNNs are capable of uncovering the implicit relations between samples, e.g., two images containing the same object, they can be applied to more general domains like vision and language.

To tackle the aforementioned problems, deep graph structure learning (GSL) is a promising solution that constructs and improves the graph topology with GNNs \cite{zhu2021deep,franceschi2019learning,chen2020iterative,jin2020graph}. Concretely, these methods parameterize the adjacency matrix with {a} probabilistic model \cite{franceschi2019learning,wang2021graph}, full parameterization \cite{jin2020graph} or metric learning model \cite{chen2020iterative,yu2020graph,fatemi2021slaps}, and jointly optimize the parameters of {the} adjacency matrix and GNNs by solving a downstream task (i.e., node classification) \cite{zhu2021deep}. However, existing methods learn graph structures in a supervised scenario, which brings the following issues: 
(1) \textit{The reliance on label information.} 
In supervised GSL methods, human-annotated labels play an important role in providing supervision signal for structure improvement. Such reliance on labels limits the application of supervised GSL on more general {cases} where annotation is unavailable. 
(2) \textit{The bias of learned edge distribution.} Node classification usually follows a semi-supervised setting, where only a small fraction of nodes (e.g., $140/2708$ in Cora dataset) are under the supervision of labels. As a result, the connections among these nodes and their neighbors would receive more guidance in structure learning, while the relations between nodes far away from them are rarely discovered by GSL \cite{fatemi2021slaps}. Such imbalance leads to the bias of edge distribution, affecting the quality of the learned structures.
(3) \textit{The limitation on downstream tasks.} In existing methods, the structure is specifically learned for node classification, so it may contain more task-specific information rather than general knowledge. Consequently, the refined topology may not benefit other downstream tasks like {link prediction} or node clustering, indicating the poor generalization {ability} of the learned structures.

\newcommand{\graphicsone}{\includegraphics[width=8.4cm]}
\begin{figure}[!t]
\vspace{-3mm}
	\centering
		\subfigure[Supervised GSL {paradigm}.]{
		\graphicsone{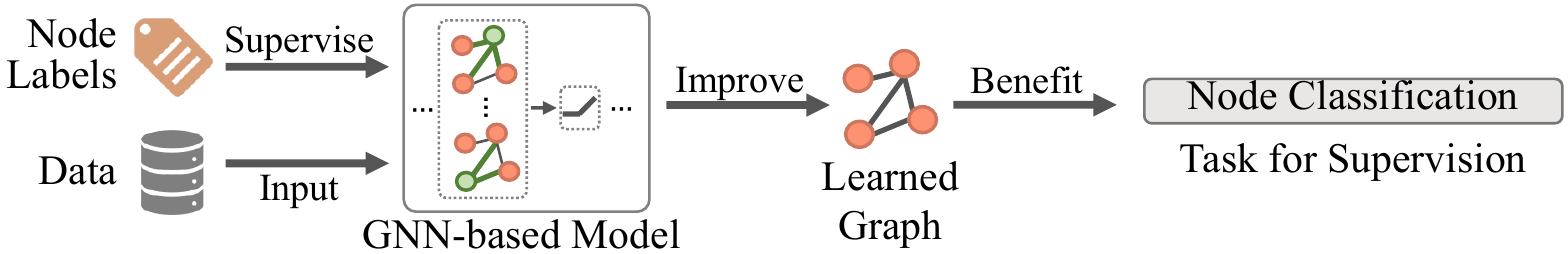}
		\label{fig:sup_gsl}
	}\\
	\vspace{-2mm}
			\subfigure[Our proposed unsupervised GSL {paradigm}.]{
		\graphicsone{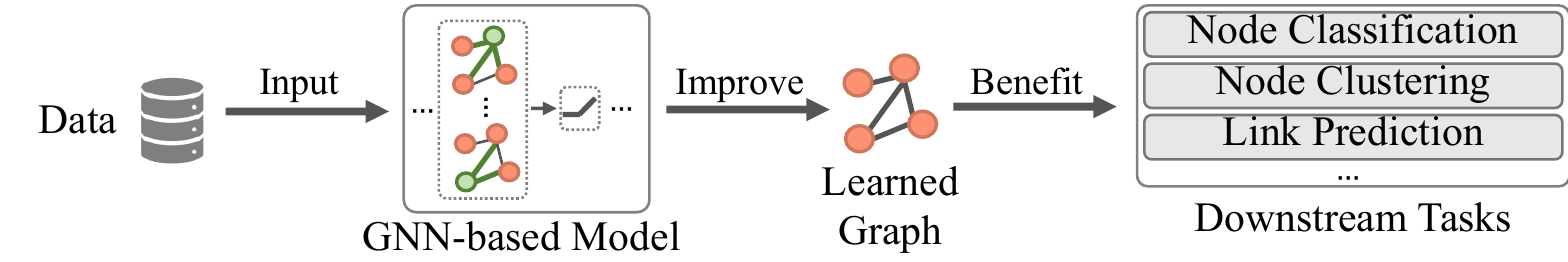}
		\label{fig:unsup_gsl}
	}
	\vspace{-4mm}
	\caption{Concept maps of (a) the existing supervised GSL {paradigm} and (b) our proposed unsupervised GSL {paradigm}.}
	\label{fig:toy_example}
	\vspace{-5mm}
\end{figure}

To address these issues, in this paper, we investigate a novel unsupervised learning {paradigm} for GSL, namely \textit{unsupervised graph structure learning}. As compared in Fig. \ref{fig:toy_example}, in our {learning paradigm}, structures are learned by data itself without any external guidance (i.e., labels), and the acquired universal, edge-unbiased topology can be freely applied to various downstream tasks. In this case, one natural question can be raised: how to provide sufficient supervision signal for unsupervised GSL? 
To answer this, we propose a novel \nameHL{S}tr\nameHL{U}cture \nameHL{B}ootstrapping contrastive \nameHL{L}earn\nameHL{I}ng fra\nameHL{ME}work (\textbf{\ourmethod} for abbreviation) to learn graph structures with the aid of self-supervised contrastive learning \cite{liu2021graph}. 
Concretely, our method constructs an ``anchor graph'' from the original data to guide structure optimization, with a contrastive loss to maximize the mutual information (MI) between anchor graph and the learned structure. 
Through maximizing their consistency, informative hidden connections can be discovered, which well respects the node proximity conveyed by the original features and structures. 
Meanwhile, as we optimize the contrastive loss on the representations of every node, all {potential} edge {candidates} will receive the essential supervision, which promotes a balanced edge distribution in the inferred topology. 
Furthermore, we design a bootstrapping mechanism to update anchor graph with the learned edges, which provides a self-enhanced supervision signal for GSL. 
Besides, we carefully design multiple graph learners and post-processing schemes to model graph topology for diverse data.
In summary, our core contributions are three-fold:
\vspace{-1mm}
\begin{itemize}
	\item \textbf{Problem.} We {propose} a novel unsupervised {learning paradigm} for graph structure learning, which is more {practical} and challenging than the existing supervised {counterpart}. To the best of our knowledge, this is the \textit{first} attempt to learn graph structures with GNNs in an unsupervised setting.
	\item \textbf{Algorithm.} We propose a novel unsupervised GSL method \ourmethod, which guides structure optimization by maximizing the agreement between the learned structure and a {crafted} self-enhanced learning target with contrastive learning.
	\item \textbf{Evaluations.} We perform extensive experiments to corroborate the effectiveness and analyze the properties of \ourmethod via thorough comparisons with state-of-the-art methods on eight benchmark datasets.
\end{itemize}
\vspace{-2mm}

\section{Related Work}
\label{sec:related_work}
\subsection{Graph Neural Networks}

Graph neural networks (GNNs) are a type of deep neural networks aiming to learn low-dimensional representations for graph-structure data \cite{kipf2017semi,wu2021comprehensive}. Modern GNNs can be categorized into two types: spectral and spatial methods. 
The spectral methods perform convolution operation to graph domain using spectral graph filter \cite{bruna2013spectral} and its simplified variants, e.g., Chebyshev polynomials filter \cite{defferrard2016convolutional} and the first-order approximation of Chebyshev filter \cite{kipf2017semi}. 
The spatial methods perform convolution operation by propagating and aggregating local information along edges in a graph \cite{hamilton2017inductive,velivckovic2018graph,xu2019how}. In spatial GNNs, different aggregation functions are designed to learn node representations, including mean/max pooling \cite{hamilton2017inductive}, LSTM \cite{hamilton2017inductive}, self-attention \cite{velivckovic2018graph}, and summation \cite{xu2019how}. Readers may refer to the elaborate survey \cite{wu2021comprehensive} for a thorough review. 
\vspace{-2mm}

\subsection{Deep Graph Structure Learning}
Graph structure learning (GSL) problem has been investigated by conventional machine learning techniques in graph signal processing \cite{egilmez2017graph}, spectral clustering \cite{bojchevski2017robust}, and network science \cite{martin2016structural}. However, these methods are not capable of handling graph data with high-dimensional features, so they are not further discussed in our paper.

Very recently, there {thrives} a branch of research {that investigates} GSL {for} GNNs {with the aim to} boost their performance on downstream tasks, which {is named} deep graph structure learning \cite{zhu2021deep}. These methods follow a general pipeline: the graph adjacency matrix is modeled with {learnable parameters}, and then jointly optimized along with GNN under the supervision of a downstream node classification task. In these methods, various techniques are leveraged to parameterize the adjacency matrix. Considering the discrete nature of graph structures, one type of methods adopts probabilistic models, such as Bernoulli probability model \cite{franceschi2019learning} and stochastic block model \cite{wang2021graph}. Another type of methods models structures with node-wise similarity computed by metric learning functions like cosine similarity \cite{chen2020iterative} and dot production \cite{yu2020graph,fatemi2021slaps}. Besides, directly treating each element in adjacency matrix as {a} learnable parameter is also an effective solution \cite{jin2020graph,fatemi2021slaps}. Nevertheless, the existing deep GSL approaches follow a supervised scenario where node labels are always required to refine the graph structures. In this paper, differently, we advocate a more {practical} unsupervised {learning paradigm} where no extra information is needed for GSL.

\vspace{-2mm}

\subsection{Contrastive Learning on Graphs}

After achieving significant performance in visual \cite{chen2020simple,grill2020bootstrap} and linguistic \cite{nlpcl1_giorgi-etal-2021-declutr,nlpcl2_chi-etal-2021-infoxlm} domains, contrastive learning has shown competitive performance and become increasingly popular in graph representation learning \cite{peng2020graph,zhu2021graph,velivckovic2019deep}. Graph contrastive learning obeys the principle of mutual information (MI) maximization, which pulls the representations of samples with shared semantic information closer while pushing the representations of irrelevant samples away \cite{liu2021graph}. In graph data, the MI maximization can be carried out to samples in the same scale (i.e., node-level \cite{zhu2021graph,jin2021multi,wan2021contrastive_nips} and graph-level \cite{you2020graph}) or different scales (i.e., node v.s. graph \cite{velivckovic2019deep,zheng2021towards} and node v.s. subgraph \cite{peng2020graph}). Graph contrastive learning also benefits diverse applications, such as chemical prediction \cite{wang2021multi}, anomaly detection \cite{liu2021anomaly,jin2021anemone}, federated learning \cite{zhang2022vertical,tan2022fedproto}, and recommendation \cite{yu2021self}. However, it still remains unclear how to effectively improve GSL using contrastive learning.

\section{Problem Definition}
\label{sec:problem_definition}
Before we make the problem statement of unsupervised GSL, we first introduce the definition of graphs.
An attributed graph can be represented by $\mathcal{G} = (\mathcal{V},\mathcal{E},\mathbf{X}) = (\mathbf{A},\mathbf{X})$, where $\mathcal{V}$ is the set of $n=|\mathcal{V}|$ nodes, $\mathcal{E}$ is the set of $m=|\mathcal{E}|$ edges, { $\mathbf{X} \in \mathbb{R}^{n \times d}$ }is the node feature matrix (where the $i$-th row $\mathbf{x}_i$ is the feature vector of node $v_i$), and $\mathbf{A} \in [0, 1]^{n \times n}$ is the weighted adjacency matrix (where $a_{ij}$ is the weight of the edge connecting $v_i$ and $v_j$). Frequently used notations are summarized in Appendix \ref{ap:notations}.

In this paper, we consider two unsupervised GSL {tasks}, i.e., structure inference and structure refinement. The former is applicable to general datasets where graph structures are not predefined or are unavailable. The latter, differently, aims to modify the given noisy topology and produce a more informative graph. Node labels are unavailable for structure optimization in both {tasks}.

\vspace{-1mm}

\begin{definition}[Structure inference] 

Given a feature matrix $\mathbf{X} \in \mathbb{R}^{n \times d}$, the target of structure inference is to automatically learn a graph topology $\mathbf S \in [0, 1]^{n \times n}$, which reflects the underlying correlations among data samples. In particular, $\mathbf S_{ij} \in [0, 1]$ indicates whether there is an edge between two samples (nodes) $\mathbf x_i$ and $\mathbf x_j$. 
\end{definition}

\begin{definition}[Structure refinement] 
Given a graph $\mathcal{G}=(\mathbf{A},\mathbf{X})$ with a noisy graph structure $\mathbf{A}$, the target of structure refinement is {to} refine {$\mathbf{A}$ to be the} optimized adjacency matrix $\mathbf S \in [0, 1]^{n \times n}$ to better capture the underlying dependency between nodes.  
\end{definition}

With the graph topology $\mathbf S$ which is either learned automatically from data or refined from an existing graph structure, the hypothesis is that the model performance on downstream tasks can be essentially improved with  $\mathcal{G}_l=(\mathbf{S},\mathbf{X})$ as the input.

\section{Methodology}
\label{sec:methodology}
\begin{figure*}[!ht]
  \includegraphics[width=0.95\textwidth]{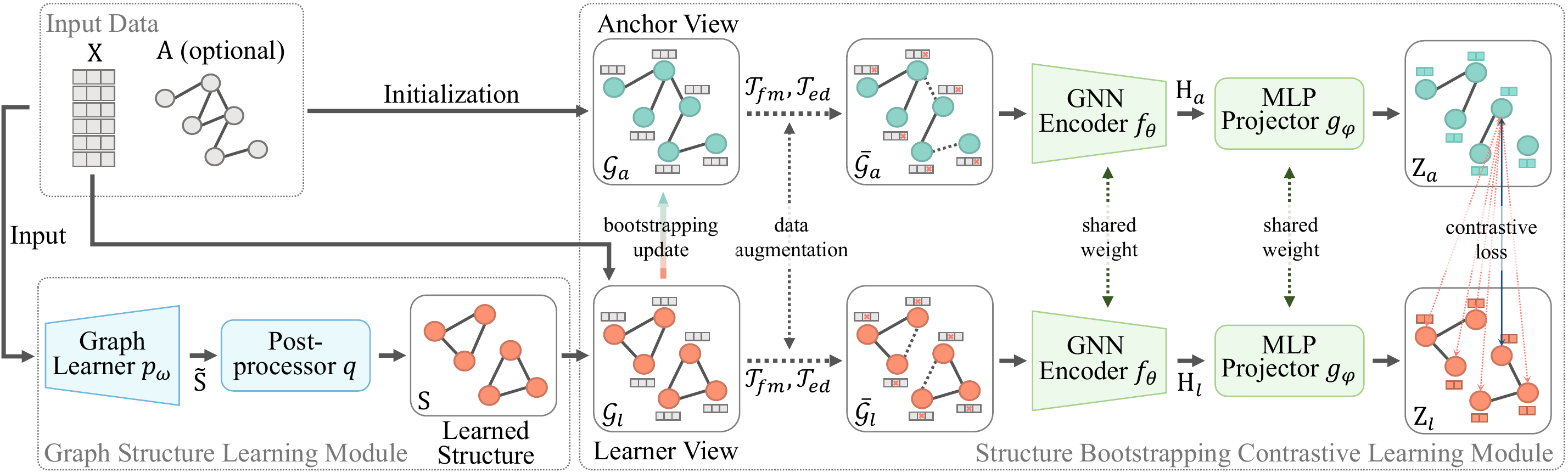}
  \vspace{-3mm}
  \caption{The overall pipeline of \ourmethod. In the graph structure learning module, the graph learner $p_\omega$ generates the sketched adjacency matrix $\tilde{\mathbf{S}}$, and then the post processor $q$ converts $\tilde{\mathbf{S}}$ into the learned structure $\mathbf{S}$. After that, the structure bootstrapping contrastive learning module optimizes $\mathbf{S}$ by maximizing the agreement between the learner view and anchor view.}
  \label{fig:pipeline}
  \vspace{-3mm}
\end{figure*}

This section elaborates our proposed \ourmethod, a novel unsupervised GSL framework. 
As shown in Fig. \ref{fig:pipeline}, \ourmethod on the highest level consists of two components: the \textit{graph structure learning module} that models and regularizes the learned graph topology and the \textit{structure bootstrapping contrastive learning module} that provides a self-optimized supervision signal for GSL. 
In the {graph structure learning module}, a sketched adjacency matrix is first parameterized by a graph learner, and then refined by a post-processor to be the learned adjacency matrix. 
Afterwards, in the {structure bootstrapping contrastive learning module}, we first establish two different views to contrast: learner view that discovers graph structure and anchor view that provides guidance for structure learning. Then, after data augmentation, the agreement between two views is maximized by a node-level contrastive learning. Specially, we design a structure bootstrapping mechanism to update anchor view with learned {structures}. 
The following subsections illustrate these crucial components respectively.

\vspace{-2mm}
\subsection{Graph Learner}

As a key component of GSL, the graph learner generates a sketched adjacency matrix $\tilde{\mathbf{S}} \in \mathbb{R}^{n \times n}$ with a parameterized model. Most existing methods \cite{franceschi2019learning,jin2020graph,chen2020iterative} adopt a single strategy to model graph structure, which cannot adapt to data with different unique properties. To find optimal structures for various data, we consider four types of graph learners, including a full graph parameterization (FGP) learner and three metric learning-based learners (i.e., Attentive, MLP, and GNN learner). 
In general, we formulate a graph learner as $p_\omega(\cdot)$, where $\omega$ is the learnable parameters. 

\textbf{FGP learner} directly models each element of the adjacency matrix by an independent parameter \cite{fatemi2021slaps,franceschi2019learning,jin2020graph} without any extra input. Formally, FGP learner is defined as:

\vspace{-2mm}

\begin{equation}
\label{eq:fgp_learner}
\tilde{\mathbf{S}}  = p_\omega^{FGP} = \sigma(\mathbf{\Omega}),
\end{equation}

\noindent where $\omega = \mathbf{\Omega} \in \mathbb{R}^{n \times n}$ is a parameter matrix and $\sigma(\cdot)$ is a non-linear function that makes training more stable. The assumption behind FGP learner is that each edge exists independently in the graph.

Different from the FGP learner, metric learning-based learners \cite{zhu2021deep,chen2020iterative} first acquire node embeddings $\mathbf{E} \in \mathbb{R}^{n \times d}$ from the input data, and then model $\tilde{\mathbf{S}}$ with pair-wise similarity of the node embeddings: 

\vspace{-2mm}

\begin{equation}
\label{eq:metric_learner}
\tilde{\mathbf{S}}  = p_\omega^{ML}(\mathbf{X},\mathbf{A}) = \phi(h_\omega(\mathbf{X},\mathbf{A})) = \phi(\mathbf{E}),
\end{equation}

\vspace{-1mm}

\noindent where $h_\omega(\cdot)$ is a neural network-based embedding function (a.k.a. embedding network) with parameter $\omega$, and $\phi(\cdot)$ is a non-parametric metric function (e.g., cosine similarity or Minkowski distance) that calculates pair-wise similarity. For different $h_\omega(\cdot)$, we provide three specific instances of metric learning-based learners: Attentive, MLP, and GNN learners.

\textbf{Attentive Learner} employs a GAT-like \cite{velivckovic2018graph} attentive network as its embedding network, where each layer compute the Hadamard production of input feature vector and the parameter vector:

\vspace{-2mm}

\begin{equation}
\label{eq:attentive_learner}
\mathbf{E}^{(l)}  = h_w^{(l)}(\mathbf{E}^{(l-1)}) = \sigma({[\mathbf{e}^{(l-1)}_1 \odot \mathbf{\omega}^{(l)}, \cdots, \mathbf{e}^{(l-1)}_n \odot \mathbf{\omega}^{(l)}]}^\intercal),
\end{equation}

\vspace{-1mm}

\noindent in which $\mathbf{E}^{(l)}$ is the output matrix of the $l$-th layer of embedding network, $\mathbf{e}^{(l-1)}_i \in \mathbb{R}^{d}$ is the transpose of the $i$-th row vector of $\mathbf{E}^{(l-1)}$, $\mathbf{\omega}^{(l)} \in \mathbb{R}^{d}$ is the parameter vector of the $l$-th layer, $\odot$ is the Hadamard operation, $(\cdot)^\intercal$ is the transposition operation, and $\sigma(\cdot)$ is a non-linear operation. The input of the first layer $\mathbf{E}^{(0)}$ is the feature matrix $\mathbf{X}$, and the output of the final layer $\mathbf{E}^{(L)}$ ($L$ is the layer number of embedding network) is the embedding matrix $\mathbf{E}$. Attentive learner assumes that each feature has different contribution to the existence of edge, but there is no significant correlation between features. 

\textbf{MLP Learner} uses a Multi-Layer Perception (MLP) as its embedding network, where a single layer can be written by:

\vspace{-2mm}

\begin{equation}
\label{eq:mlp_learner}
\mathbf{E}^{(l)}  = h_w^{(l)}(\mathbf{E}^{(l-1)}) = \sigma(\mathbf{E}^{(l-1)} \mathbf{\Omega}^{(l)}),
\end{equation}

\vspace{-1mm}

\noindent where $\mathbf{\Omega}^{(l)} \in \mathbb{R}^{d \times d}$ is the parameter martix of the $l$-th layer, and the other notations are similar to Eq. (\ref{eq:attentive_learner}). Compared to attentive learner, MLP learner further considers the correlation and combination of features, generating more informative embeddings for downstream similarity metric learning.

\textbf{GNN Learner} integrates features $\mathbf{X}$ and original structure $\mathbf{A}$ into node embeddings $\mathbf{E}$ via GNN-based embedding network. Due to the reliance on original topology, GNN learner is only {used for the} structure refinement {task}. For simplicity, we take GCN layers \cite{kipf2017semi} to form embedded network:

\vspace{-2mm}

\begin{equation}
\label{eq:gnn_learner}
\mathbf{E}^{(l)}  = h_w^{(l)}(\mathbf{E}^{(l-1)},\mathbf{A}) =  \sigma\left(\widetilde{{\mathbf{D}}}^{-\frac{1}{2}} \widetilde{{\mathbf{A}}} \widetilde{{\mathbf{D}}}^{-\frac{1}{2}} \mathbf{E}^{(l-1)} \mathbf{\Omega}^{(l)} \right),
\end{equation}

\vspace{-1mm}

\noindent where $\widetilde{{\mathbf{A}}} = \mathbf{A} + \mathbf{I}$ is the adjacency matrix with self-loop, $\widetilde{\mathbf{D}}$ is the degree matrix of $\widetilde{\mathbf{A}}$, and the other notations are similar to Eq. (\ref{eq:mlp_learner}). 
GNN Learner assumes that the connection between two nodes is related to not only features but also the original structure.

In \ourmethod, we choose the most suitable learner to model $\tilde{\mathbf{S}}$ according to the characteristics of different datasets. In Appendix \ref{ap:learners}, we analyze the properties of different graph learners and discuss how we allocate learners for each dataset.

\subsection{Post-processor}

The post-processor $q(\cdot)$ aims to refine the sketched adjacency matrix $\tilde{\mathbf{S}}$ into a sparse, non-negative, symmetric and normalized adjacency matrix $\mathbf{S}$. 
To this end, four post-processing steps are applied sequentially, i.e., sparsification $q_{sp}(\cdot)$, activation $q_{act}(\cdot)$, symmetrization $q_{sym}(\cdot)$, and normalization $q_{norm}(\cdot)$.

\noindent \textbf{Sparsification.} The sketched adjacency matrix $\tilde{\mathbf{S}}$ is often dense, representing a fully connected graph structure. However, such adjacency matrix usually makes little sense for most applications and results in expensive computation cost \cite{wang2021graph}. Hence, we conduct a k-nearest neighbors (kNN){-based} sparsification on $\tilde{\mathbf{S}}$. Concretely, for each node, we keep the edges with top-k connection values and set the rest to $0$. The sparsification $q_{sp}(\cdot)$ is expressed as:
\vspace{-1mm}
\begin{equation}
\label{eq:sparsification}
\tilde{\mathbf{S}}^{(sp)}_{ij}= q_{sp}\left(\tilde{\mathbf{S}}_{ij}\right) =\left\{\begin{aligned}
&\tilde{\mathbf{S}}_{ij},  \quad   & \tilde{\mathbf{S}}_{ij} \in \operatorname{top-k} (\tilde{\mathbf{S}}_{i}), \\
&0,   \quad   & \tilde{\mathbf{S}}_{ij} \notin \operatorname{top-k} (\tilde{\mathbf{S}}_{i}),
\end{aligned}\right.
\vspace{-1mm}
\end{equation}
\noindent where $\operatorname{top-k} (\tilde{\mathbf{S}}_{i})$ is the set of top-k values of row vector $\tilde{\mathbf{S}}_{i}$. To keep the gradient flow, we do not apply sparsification for the FGP learner. For large-scale graphs, we perform the kNN sparsification with its locality-sensitive approximation \cite{fatemi2021slaps} where the nearest neighbors are selected from a batch of nodes instead of all nodes, which reduces the requirement of memory.

\noindent \textbf{Symmetrization and Activation.} In real-world graphs, the connections are often bi-directional, which requires a symmetric adjacency matrix. In addition, the edge weights should be non-negative according to the definition of adjacency matrix. To meet these conditions, the symmetrization and activation are performed as:
\vspace{-1mm}
\begin{equation}
\label{eq:sym_act}
\tilde{\mathbf{S}}^{(sym)} = q_{sym}\left(q_{act}\left(\tilde{\mathbf{S}}^{(sp)}\right)\right) = \frac{\sigma_q\left(\tilde{\mathbf{S}}^{(sp)}\right) + \sigma_q\left(\tilde{\mathbf{S}}^{(sp)}\right)^\intercal}{2},
\vspace{-1mm}
\end{equation}

\noindent where $\sigma_q(\cdot)$ is a non-linear activation. For metric learning-based learners, we define $\sigma_q(\cdot)$ as ReLU function. For FGP learner, we apply the ELU function to prevent gradient from disappearing.

\noindent \textbf{Normalization.} To guarantee the edge weights are within the range $[0, 1]$, we finally conduct a normalization on $\tilde{\mathbf{S}}$. In particular, we apply a symmetrical normalization: 
\vspace{-1mm}
\begin{equation}
\label{eq:normalization}
\mathbf{S} = q_{norm}\left(\tilde{\mathbf{S}}^{(sym)}\right) = \left(\tilde{\mathbf{D}}^{(sym)}\right)^{-\frac{1}{2}} \tilde{\mathbf{S}}^{(sym)} \left(\tilde{\mathbf{D}}^{(sym)}\right)^{-\frac{1}{2}},
\vspace{-1mm}
\end{equation}

\noindent where $\tilde{\mathbf{D}}^{(sym)}$ is the degree matrix of $\tilde{\mathbf{S}}^{(sym)}$. 

\subsection{Multi-view Graph Contrastive Learning}

Since we have obtained a well-parameterized adjacency matrix $\mathbf{S}$, a natural question that arises here is: \textit{how to provide an effective supervision signal guiding the graph structure learning without label information?} Our answer is to acquire the supervision signal from data itself via multi-view graph contrastive learning. To be concrete, we construct two graph views based on the learned structure and the original data respectively. Then, data augmentation is applied to both views. Finally, we maximize the MI between two augmented views with node-level contrastive learning. 

\vspace{-2mm}

\subsubsection{Graph View Establishment}

Different from general graph contrastive learning methods \cite{zhu2021graph,jin2021multi} that obtain both views from the original data, \ourmethod defines the learned graph as one view, and constructs the other view with input data. The former, named \textit{learner view}, explores potential structures in every step. The latter, named \textit{anchor view}, provides a stable {learning} target for GSL. 

\textbf{Learner view} is directly built by integrating the learned adjacency matrix $\mathbf{S}$ and the feature matrix $\mathbf{X}$ together, which is denoted as $\mathcal{G}_l=(\mathbf{S},\mathbf{X})$. In each training iteration, $\mathbf{S}$ and the parameters used to model it are directly updated by gradient descent to discover optimal graph structures. In \ourmethod, we initialize learner views as the kNN graph built on features, since it is an effective way to provide a starting point for GSL, as suggested in \cite{franceschi2019learning,fatemi2021slaps}. Specifically, for FGP learner, we initialize the parameters corresponding to kNN edges as $1$ while the rest as $0$. For attentive learner, we let each element in $\omega^{(l)} \in \mathcal{\omega}$ to be $1$. Then, feature-level similarities are computed according to the metric function, and the kNN graph is obtained by the sparsification post-processing. For MLP and GNN learners, similarly, we set the embedding dimension to be $d$ and initialize $\Omega^{(l)} \in \mathcal{\omega}$ as identity matrices.

\textbf{Anchor view} plays a ``teacher'' role that provides correct and stable guidance for GSL. For the structure refinement {task} where the original structure $\mathbf{A}$ is available, we define anchor view as $\mathcal{G}_a=(\mathbf{A}_{a},\mathbf{X})=(\mathbf{A},\mathbf{X})$; for the structure inference {task} where $\mathbf{A}$ is inaccessible, we take an identity matrix $\mathbf{I}$ as the anchor structure: $\mathcal{G}_a=(\mathbf{A}_{a},\mathbf{X})=(\mathbf{I},\mathbf{X})$. To provide a stable {learning} target, anchor view is not updated by gradient descent but a novel bootstrapping mechanism which will be introduced in Section \ref{subsec:bootstrap_mechanism}. 

\vspace{-2mm}

\subsubsection{Data Augmentation}

In contrastive learning, data augmentation is a key to {benefiting} the model {through exploring} richer underlying semantic information by making the learning tasks more challenging to solve  \cite{chen2020simple,zhu2021graph,liu2021anomaly}. In \ourmethod, we exploit two simple but effective augmentation schemes, i.e., \textit{feature masking} and \textit{edge dropping}, to corrupt the graphs views at both structure and feature levels.

\noindent \textbf{Feature masking.} To disturb the node features, we randomly select a fraction of feature dimensions and mask them with zeros. Formally, for a given feature matrix $\mathbf{X}$, a masking vector $\mathbf{m}^{(x)} \in \{0,1\}^{d}$ is first sampled, where each element is drawn from a Bernoulli distribution with probability $p^{(x)}$ independently. Then, we mask the feature vector of each node with $\mathbf{m}^{(x)}$:

\vspace{-1mm}

\begin{equation}
\label{eq:maskfeat}
\overline{\mathbf{X}} = \mathcal{T}_{fm}(\mathbf{X}) = {[\mathbf{x}_1 \odot \mathbf{m}^{(x)}, \cdots, \mathbf{x}_n \odot \mathbf{m}^{(x)}]}^\intercal,
\vspace{-1mm}
\end{equation}

\noindent where $\overline{\mathbf{X}}$ is the augmented feature matrix, $\mathcal{T}_{fm}(\cdot)$ is the feature masking transformation, and $\mathbf{x}_i$ is the transpose of the i-th row vector of $\mathbf{X}$. 

\noindent \textbf{Edge dropping.} Apart from masking features, we corrupt the graph structure by randomly dropping a portion of edges. Specifically, for a given adjacency matrix $\mathbf{A}$, we first sample a masking matrix $\mathbf{M}^{(a)} \in \{0,1\}^{n \times n}$, where each element $\mathbf{M}^{(a)}_{ij}$ is drawn from a Bernoulli distribution with probability $p^{(a)}$ independently. After that, the adjacency matrix is masked with $\mathbf{M}^{(a)}$: 

\vspace{-1mm}

\begin{equation}
\label{eq:dropedge}
\overline{\mathbf{A}} = \mathcal{T}_{ed}(\mathbf{A}) = \mathbf{A} \odot \mathbf{M}^{(a)},
\vspace{-1mm}
\end{equation}

\noindent where $\overline{\mathbf{A}}$ is the augmented adjacency matrix, and $\mathcal{T}_{ed}(\cdot)$ is the edge dropping transformation.

In \ourmethod, we jointly leverage these two augmentation schemes to generate augmented graphs on both learner and anchor views:

\vspace{-1mm}

\begin{equation}
\label{eq:augmentation}
\overline{\mathcal{G}}_l = (\mathcal{T}_{ed}(\mathbf{S}), \mathcal{T}_{fm}(\mathbf{X})),\; 
\overline{\mathcal{G}}_a = (\mathcal{T}_{ed}(\mathbf{A}_a), \mathcal{T}_{fm}(\mathbf{X})) ,
\vspace{-0.3mm}
\end{equation}

\noindent where $\overline{\mathcal{G}}_l$ and $\overline{\mathcal{G}}_a$ are the augmented learner view and anchor view, respectively. To obtain different contexts in the two views, the feature masking for two views employs different probabilities $p^{(x)}_l \neq p^{(x)}_a$. For edge dropping, since the adjacency matrices of two views are already significantly different, we use the same dropping probability $p^{(a)}_l = p^{(a)}_a = p^{(a)}$. Note that other advanced augmentation schemes can also be applied to \ourmethod, which is left for our future research.

\vspace{-2mm}

\subsubsection{Node-level Contrastive Learning}

After obtaining two augmented graph views, we perform a node-level contrastive learning to maximize the MI between them. In \ourmethod, we adopt a simple contrastive learning framework originated from SimCLR \cite{chen2020simple} which consists of the following components:

\noindent \textbf{GNN-based encoder.} A GNN-based encoder $f_\theta(\cdot)$ extracts node-level representations for augmented graphs $\overline{\mathcal{G}}_l$ and $\overline{\mathcal{G}}_a$:   
\vspace{-1mm}

\begin{equation}
\label{eq:encoding}
{\mathbf{H}}_{l} = f_\theta(\overline{\mathcal{G}}_{l}),\; {\mathbf{H}}_{a} = f_\theta(\overline{\mathcal{G}}_{a}),
\vspace{-1mm}
\end{equation}

\noindent where $\theta$ is the parameter of encoder $f_\theta(\cdot)$, and ${\mathbf{H}}_{l}$, ${\mathbf{H}}_{a} \in \mathbb{R}^{n \times d_1}$ ($d_1$ is the representation dimension) are the node representation matrices for learner/anchor views, respectively. In \ourmethod, we utilize GCN \cite{kipf2017semi} as our encoder and set its layer number $L_1$ to $2$.

\noindent \textbf{MLP-based projector.} Following the encoder, a projector $g_\varphi(\cdot)$ with $L_2$ MLP layers maps the representations to another latent space where the contrastive loss is calculated: 
\vspace{-1mm}

\begin{equation}
\label{eq:projection}
{\mathbf{Z}}_{l} = g_\varphi({\mathbf{H}}_{l}),\; {\mathbf{Z}}_{a} = g_\varphi({\mathbf{H}}_{a}),
\vspace{-1mm}
\end{equation}

\noindent where $\varphi$ is the parameter of projector $g_\varphi(\cdot)$, and ${\mathbf{Z}}_{l}$, ${\mathbf{Z}}_{a} \in \mathbb{R}^{n \times d_2}$ ($d_2$ is the projection dimension) are the {projected node representation} matrices for learner/anchor views, respectively. 

\noindent \textbf{Node-level contrastive loss function.} A contrastive loss $\mathcal{L}$ is leveraged to enforce maximizing the agreement between the projections  $z_{l,i}$ and  $z_{a,i}$ of the same node $v_i$ on two views. In our framework, a symmetric normalized temperature-scaled cross-entropy loss (NT-Xent) \cite{oord2018representation,sohn2016improved} is applied:

\vspace{-1mm}

\begin{equation}
\label{eq:cl_loss}
\begin{aligned}
\mathcal{L} = \frac{1}{2n} \sum_{i=1}^{n} \Big[ \ell(z_{l,i},z_{a,i}) + \ell(z_{a,i},z_{l,i}) \Big], \\
\ell(z_{l,i},z_{a,i}) = \operatorname{log}\frac{e^{\operatorname{sim}(\mathbf{z}_{l,i},\mathbf{z}_{a,i})/t}}{\sum_{k=1}^{n} e^{\operatorname{sim}(\mathbf{z}_{l,i},\mathbf{z}_{a,k})/t} },
\end{aligned}
\vspace{-0.5mm}
\end{equation}

\noindent where $\operatorname{sim}(\cdot,\cdot)$ is the cosine similarity function, and $t$ is the temperature parameter. $\ell(z_{a,i},z_{l,i})$ {is} computed {following} $\ell(z_{l,i},z_{a,i})$.

\subsection{Structure Bootstrapping Mechanism} \label{subsec:bootstrap_mechanism}

With a fixed anchor adjacency matrix $\mathbf{A}_a$ defined by $\mathbf{A}$ or $\mathbf{I}$, \ourmethod can learn graph structure $\mathbf{S}$ by maximizing the MI between two views. 
However, using a constant anchor graph may lead to several issues: 
(1) \textit{Inheritance of error information}. Since $\mathbf{A}_a$ is directly borrowed from the input data, it would carry some natural noise (e.g., missing or redundant edges) of the original graph. If the noise is not eliminated in the learning process, the learned structures will finally inherit it. 
(2) \textit{Lack of persistent guidance}. A fixed anchor graph contains limited information to guide GSL. Once the graph learner captures this information, it will be hard for the model to gain effective supervision in the following training steps. 
(3) \textit{Over-fitting the anchor structure.} Driven by the learning objective that maximizes the agreement between two views, the learned structure tends to over-fit the fixed anchor structure, resulting in a similar testing performance to the original data.

Inspired by previous bootstrapping-based algorithms \cite{caron2018deep,tarvainen2017mean,grill2020bootstrap}, we design a structure bootstrapping mechanism to provide a self-enhanced anchor view as the learning target. The core idea of our solution is to update the anchor structure $\mathbf{A}_a$ with a slow-moving {augmentation} of the learned structure $\mathbf{S}$ instead of keeping $\mathbf{A}_a$ unchanged. In particular, given a decay rate $\tau \in [ 0,1 ]$, the anchor structure $\mathbf{A}_a$ is updated every $c$ iterations as following:

\vspace{-1mm}
\begin{equation}
\label{eq:momentum_update}
\mathbf{A}_a \gets \tau \mathbf{A}_a + (1 - \tau) \mathbf{S}.
\vspace{-0.5mm}
\end{equation}

\begin{table*}[htbp!]
\vspace{-2mm}
\centering
\caption{Node classification accuracy (percentage with standard deviation) in structure inference scenario. Available data for \textit{graph structure learning} during the training phase is shown in the first column, where $\mathbf{X}$, $\mathbf{Y}$, $\mathbf{A}_{knn}$ correspond to node features, labels and the adjacency matrix of kNN graph, respectively. The highest and second highest results are highlighted with \textcolor{MyRed}{boldface} and \textcolor{red}{\underline{underline}}, respectively. The symbol ``OOM'' means out of memory.} 
\vspace{-3mm}
\begin{adjustbox}{width=2\columnwidth,center}
	\begin{tabular}{ll|p{1.6cm}<{\centering} p{1.6cm}<{\centering} p{1.6cm}<{\centering} p{1.6cm}<{\centering} p{1.6cm}<{\centering} p{1.6cm}<{\centering} p{1.6cm}<{\centering} p{1.6cm}<{\centering} }
	\toprule
	\multirow{2}*{\textbf{\tabincell{c}{Available \\Data for GSL }}} & \multirow{2}*{\textbf{Method}} & \multicolumn{8}{c} {\textbf{Dataset}} \\
	\cline{3-10}
	&  & Cora & Citeseer & Pubmed & ogbn-arxiv & Wine & Cancer & Digits & 20news\\
	\hline
	- & LR & 60.8$\pm$0.0	 & 62.2$\pm$0.0 & {72.4$\pm$0.0} & 52.5$\pm$0.0 & 92.1$\pm$1.3 & 93.3$\pm$0.5 & 85.5$\pm$1.5 & 42.7$\pm$1.7\\
	- & Linear SVM & 58.9$\pm$0.0  & 58.3$\pm$0.0 & 72.7$\pm$0.1 & 51.8$\pm$0.0 & 93.9$\pm$1.6 & 90.6$\pm$4.5 & 87.1$\pm$1.8 & 40.3$\pm$1.4\\
	- & MLP        & 56.1$\pm$1.6 & 56.7$\pm$1.7 & 71.4$\pm$0.0 & 54.7$\pm$0.1 & 89.7$\pm$1.9 & 92.9$\pm$1.2 & 36.3$\pm$0.3 & 38.6$\pm$1.4\\
	\hline
	- & GCN$_{knn}$ \cite{kipf2017semi}       & 66.5$\pm$0.4 & 68.3$\pm$1.3 & 70.4$\pm$0.4 & 54.1$\pm$0.3 & 93.2$\pm$3.1 & 83.8$\pm$1.4 & 91.3$\pm$0.5 & 41.3$\pm$0.6\\
	- & GAT$_{knn}$ \cite{velivckovic2018graph}       & 66.2$\pm$0.5 & 70.0$\pm$0.6	& 69.6$\pm$0.5 & OOM & 91.5$\pm$2.4 & 95.1$\pm$0.8 & 91.4$\pm$0.1	& 45.0$\pm$1.2\\
	- & SAGE$_{knn}$ \cite{hamilton2017inductive}  & 66.1$\pm$0.7 & 68.0$\pm$1.6 & 68.7$\pm$0.2	& 55.2$\pm$0.4 & 87.4$\pm$0.8 & 93.7$\pm$0.3 & 91.6$\pm$0.7 & 45.4$\pm$0.4\\
	\hline
	\xy   & LDS  \cite{franceschi2019learning}      & 71.5$\pm$0.8 & 71.5$\pm$1.1	& OOM & OOM	& 97.3$\pm$0.4 & 94.4$\pm$1.9 & 92.5$\pm$0.7 & 46.4$\pm$1.6\\
	\xyak & GRCN  \cite{yu2020graph}     & 69.6$\pm$0.2 & 70.4$\pm$0.3 & 70.6$\pm$0.1	& OOM & 96.6$\pm$0.4 & 95.4$\pm$0.6 & 92.8$\pm$0.2	& 41.8$\pm$0.2\\
	\xyak & Pro-GNN \cite{jin2020graph}   & 69.2$\pm$1.4 & 69.8$\pm$1.7	& OOM & OOM	& 95.1$\pm$1.5 & 96.5$\pm$0.1 & 93.9$\pm$1.9 & 45.7$\pm$1.4\\
	\xyak & GEN  \cite{wang2021graph}      & 69.1$\pm$0.7 & 70.7$\pm$1.1 & 70.7$\pm$0.9 & OOM & 96.9$\pm$1.0 & \secHL{96.8$\pm$0.4} & 94.1$\pm$0.4	& 47.1$\pm$0.3\\
	\xy   & IDGL \cite{chen2020iterative}      & 70.9$\pm$0.6 & 68.2$\pm$0.6 & 70.1$\pm$1.3 & 55.0$\pm$0.2 & \secHL{98.1$\pm$1.1} & 95.1$\pm$1.0 & 93.2$\pm$0.9 & 48.5$\pm$0.6\\
	\xy   & SLAPS	\cite{fatemi2021slaps}     & \fstHL{73.4$\pm$0.3}	& \secHL{72.6$\pm$0.6}	& \fstHL{74.4$\pm$0.6}	& \fstHL{56.6$\pm$0.1}	& 96.6$\pm$0.4	& 96.6$\pm$0.2	& \fstHL{94.4$\pm$0.7}	& \fstHL{50.4$\pm$0.7} \\
	\hline
	\ak   & GDC   \cite{klicpera2019diffusion}     &68.1$\pm$1.2&	68.8$\pm$0.8&	68.4$\pm$0.4&	OOM	&96.1$\pm$1.0&	95.9$\pm$0.4&	92.6$\pm$0.5&	46.4$\pm$0.9\\
	\onlyx& SLAPS-2s \cite{fatemi2021slaps} & 72.1$\pm$0.4	& 69.4$\pm$1.4	& 71.1$\pm$0.5	& 54.2$\pm$0.2	& 96.2$\pm$2.1	& 95.9$\pm$1.2	& 93.6$\pm$0.8	& 47.7$\pm$0.7\\
	\onlyx& \textbf{\ourmethod}	& \secHL{73.0$\pm$0.6}	& \fstHL{73.1$\pm$0.3}	& \secHL{73.8$\pm$0.6}	& \secHL{55.5$\pm$0.1}	& \fstHL{98.2$\pm$1.6}	& \fstHL{97.2$\pm$0.2}	& \secHL{94.3$\pm$0.4}	& \secHL{49.2$\pm$0.6}\\
	\bottomrule
	
	\end{tabular}
	\end{adjustbox}
	\label{tab:cls_si}
	\vspace{-1mm}
\end{table*}

Benefiting from the structure bootstrapping mechanism, \ourmethod has nice properties that can address the aforementioned problems. 
With the process of updating, the weights of some noise edges gradually decrease in $\mathbf{A}_a$, which relieves {their negative} impact on {structure learning}.
Meanwhile, since the learning target $\mathbf{A}_a$ is changing during the training phase, it can always {incorporate more} effective information to guide the learning of topology, and the over-fitting problem is naturally resolved. 
More importantly, our structure bootstrapping mechanism leverages the learned knowledge to improve the learning target in turn, pushing the model to discover {increasingly} optimal graph structure {constantly}. 
Besides, the slow-moving average (with $\tau > 0.99$) updating ensures the stability of training.

\subsection{Overall Framework}

In this subsection, we first illustrate the training process of \ourmethod, and then introduce the tricks to help apply it to large-scale graphs. 

\noindent \textbf{Model training.} In our training process, we first initialize the parameters and anchor adjacency matrix $\mathbf{A}_a$. Then, in each iteration, we perform forward propagation to compute the contrastive loss $\mathcal{L}$, and update all the parameters jointly via back propagation. After back propagation, we update $\mathbf{A}_a$ by bootstrapping structure mechanism every $c$ iterations. Finally, we acquire the learned topology represented by $\mathbf{S}$. As analyzed in Appendix \ref{ap:complexity}, the time complexity of \ourmethod is $\mathcal{O}(n^2d+md_1L_1 + nd_1^2L_1 + nd_2^2L_2 + nk)$. The {algorithmic description} is provided in Appendix \ref{ap:algo}.

\noindent \textbf{Scalability extension.} To extend the scalability of \ourmethod, the key is to avoid $\mathcal{O}(n^2)$ space complexity and time complexity. To this end, we adopt the following measures: (1) To avoid explosive number of parameters, we use metric learning-based learners instead of FGP learner. (2) For sparsification post-processing, we consider a locality-sensitive approximation for kNN graph \cite{fatemi2021slaps}. (3) For graph contrastive learning, we compute the contrastive loss $\mathcal{L}$ for a mini-batch of samples instead of all nodes. (4) To reduce the space complexity of the bootstrapped structure, we perform the update in Eq. (\ref{eq:momentum_update}) with a larger iteration interval $c$ ($c\geq10$).

\section{Experiments}
\label{sec:experiments}
In this section, we conduct empirical experiments to demonstrate the effectiveness of the proposed framework \ourmethod. We aim to answer five research questions as follows:
\textbf{RQ1:} How effective is \ourmethod for learning graph structure under unsupervised settings? 
\textbf{RQ2:} How does the structure bootstrapping mechanism influence the performance of \ourmethod? 
\textbf{RQ3:} How do key hyper-parameters impact the performance of \ourmethod? 
\textbf{RQ4:} How robust is \ourmethod to adversarial graph structures? 
and \textbf{RQ5:} What kind of graph structure is learned by \ourmethod?

\vspace{-3mm}

\subsection{Experimental Setups}

\textbf{Downstream tasks for evaluation.} We use node classification and node clustering tasks to evaluate the quality of learned topology. For node classification, We conduct experiments on both structure inference/refinement scenarios, and use classification accuracy as our metric. For node clustering, the experiments are conducted on structure refinement scenario, and four metrics are employed, including clustering accuracy (C-ACC), Normalized Mutual Information (NMI), F1-score (F1) and Adjusted Rand Index (ARI).

\noindent \textbf{Datasets.} We evaluate \ourmethod on eight real-world benchmark datasets, including four graph-structured datasets (i.e., Cora, Citeseer \cite{sen2008collective}, Pubmed \cite{pubmed_namata2012query} and ogbn-arxiv \cite{hu2020ogb}) and four non-graph datasets (i.e., Wine, Cancer, Digits and 20news \cite{asuncion2007uci}). Details of datasets are summarized in Appendix \ref{app:dataset}.

\noindent \textbf{Baselines.} For node classification, we mainly compare \ourmethod with two categories of methods, including three structure-fixed GNN methods (i.e., GCN \cite{kipf2017semi}, GAT \cite{velivckovic2018graph} and GraphSAGE (SAGE for short) \cite{hamilton2017inductive}), and six supervised GSL methods (i.e., LDS \cite{franceschi2019learning}, GRCN \cite{yu2020graph}, Pro-GNN \cite{jin2020graph}, GEN \cite{wang2021graph}, IDGL \cite{chen2020iterative} and SLAPS \cite{fatemi2021slaps}). We also consider GDC \cite{klicpera2019diffusion}, a diffusion-based graph structure improvement method, and SLAPS-2s, a variant of SLAPS \cite{fatemi2021slaps} which only uses denoising autoencoder to learn topology, as two baselines of unsupervised GSL. In structure inference scenario, we further add three conventional feature-based classifiers (Logistic Regression, Linear SVM and MLP) for comparison. 
For node clustering task, we consider baseline methods belonging to the following three categories: 1) feature-based clustering methods (i.e., K-means \cite{kmeans_hartigan1979algorithm} and Spectral Clustering (SC for short) \cite{sc_ng2002spectral}); 2) structure-based clustering methods (i.e., GraphEncoder (GE for short) \cite{ge_tian2014learning}, DeepWalk (DW for short) \cite{dw_perozzi2014deepwalk}, DNGR \cite{dngr_cao2016deep} and M-NMF \cite{mnmf_wang2017community}); and 3) attributed graph clustering methods (i.e., RMSC \cite{rmsc_xia2014robust}, TADW \cite{tadw_yang2015network}, VGAE \cite{vgae_kipf2016variational}, ARGA \cite{arga_pan2018adversarially}, MGAE \cite{wang2017mgae}, AGC \cite{agc_zhang2019attributed} and DAEGC \cite{daegc_wang2019attributed}). 

For other experimental details, including infrastructures and hyper-parameter, interested readers can refer to Appendix \ref{app:exp_detail}. Our code is available at \url{https://github.com/GRAND-Lab/SUBLIME}.

\subsection{Performance Comparison (RQ1)}

\begin{table}[tbp!]
\vspace{-3mm}
\centering
\caption{Node classification accuracy (percentage with standard deviation) in structure refinement scenario.} 
\vspace{-4mm}
\begin{adjustbox}{width=1.04\columnwidth,center}
	\begin{tabular}{ll|cccc}
	\toprule
	\multirow{2}*{\textbf{\tabincell{c}{Available \\ Data for GSL}}} & \multirow{2}*{\textbf{Method}} & \multicolumn{4}{c} {\textbf{Dataset}} \\
	\cline{3-6}
	&  & Cora & Citeseer & Pubmed & ogbn-arxiv \\
	\hline
	- & GCN      &  81.5 & 70.3 & 79.0 & 71.7$\pm$0.3\\
	- & GAT      &  83.0$\pm$0.7 & 72.5$\pm$0.7 & 79.0$\pm$0.3 & OOM\\
	- & SAGE     &  77.4$\pm$1.0 & 67.0$\pm$1.0 & 76.6$\pm$0.8 & 71.5$\pm$0.3\\
	\hline
	\xya   & LDS      &  {83.9$\pm$0.6} & \fstHL{74.8$\pm$0.3} & OOM & OOM\\
	\xya   & GRCN     &  \secHL{84.0$\pm$0.2} & 73.0$\pm$0.3 & 78.9$\pm$0.2 & OOM\\
	\xya   & Pro-GNN  &  82.1$\pm$0.4 & 71.3$\pm$0.4 & OOM & OOM\\
	\xya   & GEN        &  82.3$\pm$0.4 & \secHL{73.5$\pm$1.5} & 80.9$\pm$0.8 & OOM\\
	\xya   & IDGL       &  \secHL{84.0$\pm$0.5} & {73.1$\pm$0.7} & \fstHL{83.0$\pm$0.2} & \fstHL{72.0$\pm$0.3}\\
	\hline
	\onlya & GDC        &  83.6$\pm$0.2 & 73.4$\pm$0.3 & 78.7$\pm$0.4 & OOM\\
	\xa    & \textbf{\ourmethod}	&  \fstHL{84.2$\pm$0.5}	 & \secHL{73.5$\pm$0.6} & \secHL{81.0$\pm$0.6} & \secHL{71.8$\pm$0.3}\\
	\bottomrule
	
	\end{tabular}
	\end{adjustbox}
	\label{tab:cls_sr}
	
	\vspace{-3mm}
\end{table}

\begin{table}[tbp!]
\centering
\caption{Node clustering performance (4 metrics in percentage) in structure refinement scenario.}
\vspace{-4mm}
\begin{adjustbox}{width=1.04\columnwidth,center}
	\begin{tabular}{l|cccc|cccc}
	\toprule
	\multirow{2}*{\textbf{{Method}}} & \multicolumn{4}{c} {\textbf{Cora}} & \multicolumn{4}{c} {\textbf{Citeseer}} \\
	\cline{2-5} \cline{6-9}
	& C-ACC & NMI & F1 & ARI & C-ACC & NMI & F1 & ARI \\
	\hline
	K-means & 50.0 & 31.7 & 37.6 & 23.9 & 54.4 & 31.2 & 41.3 & 28.5 \\
	SC      & 39.8 & 29.7 & 33.2 & 17.4 & 30.8 & 9.0 & 25.7 & 8.2 \\
	\hline
	GE      & 30.1 & 5.9 & 23.0 & 4.6 & 29.3 & 5.7 & 21.3 & 4.3 \\
	DW      & 52.9 & 38.4 & 43.5 & 29.1 & 39.0 & 13.1 & 30.5 & 13.7 \\
	DNGR    & 41.9 & 31.8 & 34.0 & 14.2 & 32.6 & 18.0 & 30.0 & 4.3 \\
	M-NMF   & 42.3 & 25.6 & 32.0 & 16.1 & 33.6 & 9.9 & 25.5 & 7.0 \\
	\hline
	RMSC    & 46.6 & 32.0 & 34.7 & 20.3 & 51.6 & 30.8 & 40.4 & 26.6 \\
	TADW    & 53.6 & 36.6 & 40.1 & 24.0 & 52.9 & 32.0 & 43.6 & 28.6 \\
	VGAE    & 59.2 & 40.8 & 45.6 & 34.7 & 39.2 & 16.3 & 27.8 & 10.1 \\
	ARGA    & 64.0 & 44.9 & 61.9 & 35.2 & 57.3 & 35.0 & 54.6 & 34.1 \\
	MGAE    & 68.1 & 48.9 & 53.1 & \fstHL{56.5} & 66.9 & \secHL{41.6} & 52.6 & \secHL{42.5} \\
	AGC     & 68.9 & \secHL{53.7} & \secHL{65.6} & 44.8 & 67.0 & 41.1 & 62.5 & 41.5 \\
	DAEGC   & \secHL{70.4} & 52.8 & \fstHL{68.2} & 49.6 & \secHL{67.2} & 39.7 & \fstHL{63.6} & 41.0 \\
	\hline
	\textbf{\ourmethod} & \fstHL{71.3} & \fstHL{54.2} & 63.5 & \secHL{50.3} & \fstHL{68.5} & \fstHL{44.1} & \secHL{63.2} & \fstHL{43.9} \\
	\bottomrule
	\end{tabular}
	\end{adjustbox}
	\label{tab:clu_sr}
	\vspace{-3mm}
\end{table}

\textbf{Node classification in structure inference scenario.}
Table \ref{tab:cls_si} reports the classification accuracy of our method and other baselines in structure inference scenario. For structure-fixed GNNs (i.e., GCN, GAT and GraphSAGE) and GSL methods designed for structure refinement scenarios (i.e., GRCN, Pro-GNN, GEN and GDC), we use kNN graphs as their input graphs, where $k$ is tuned in the same search space to our method. 

As can be observed, without the guidance of labels, our proposed \ourmethod outperforms all baselines on 3 out of 8 benchmarks and achieves the runner-up results on the rest datasets. This competitive performance benefits from the novel idea of guiding GSL with a self-enhanced learning target by graph contrastive learning. Besides, the result on ogbn-arxiv exhibits the scailbility of \ourmethod.  

We make other observations as follows. Firstly, the performance of structure-fixed GNNs (taking kNN graphs as input) is superior to conventional feature-based classifiers on most datasets, which shows the benefit of considering the underlying relationship among samples. Secondly, GSL methods achieve better performance than structure-fixed methods, indicating the significance of structure optimization. Thirdly, compared to supervised GSL methods, the unsupervised methods also achieve competitive results without the supervision of labels, which shows their effectiveness. 

\noindent \textbf{Node classification in structure refinement scenario.}
Table \ref{tab:cls_sr} summarizes the classification performance of each method in structure refinement scenario. 
We find that \ourmethod still shows very promising results against not only the self-supervised but also supervised methods, indicating that \ourmethod can leverage self-supervision signal to improve the original graphs effectively. 

\noindent \textbf{Node clustering in structure refinement scenario.}
In Table \ref{tab:clu_sr}, we report the results of node clustering. Compared to baselines, our performance improvement illustrates that optimizing graph structures is indeed helpful to the clustering task. Meanwhile, the implementation of \ourmethod for node clustering task suggests that our learned topology can be applied to not only node classification task but also a wide range of downstream tasks. 

\vspace{-1mm}

\subsection{Ablation Study (RQ2)}

\begin{figure}[t!]
\vspace{-3mm}
	\centering
		\subfigure[Test accuracy w.r.t. epoch.]{
		\includegraphics[height=3.5cm]{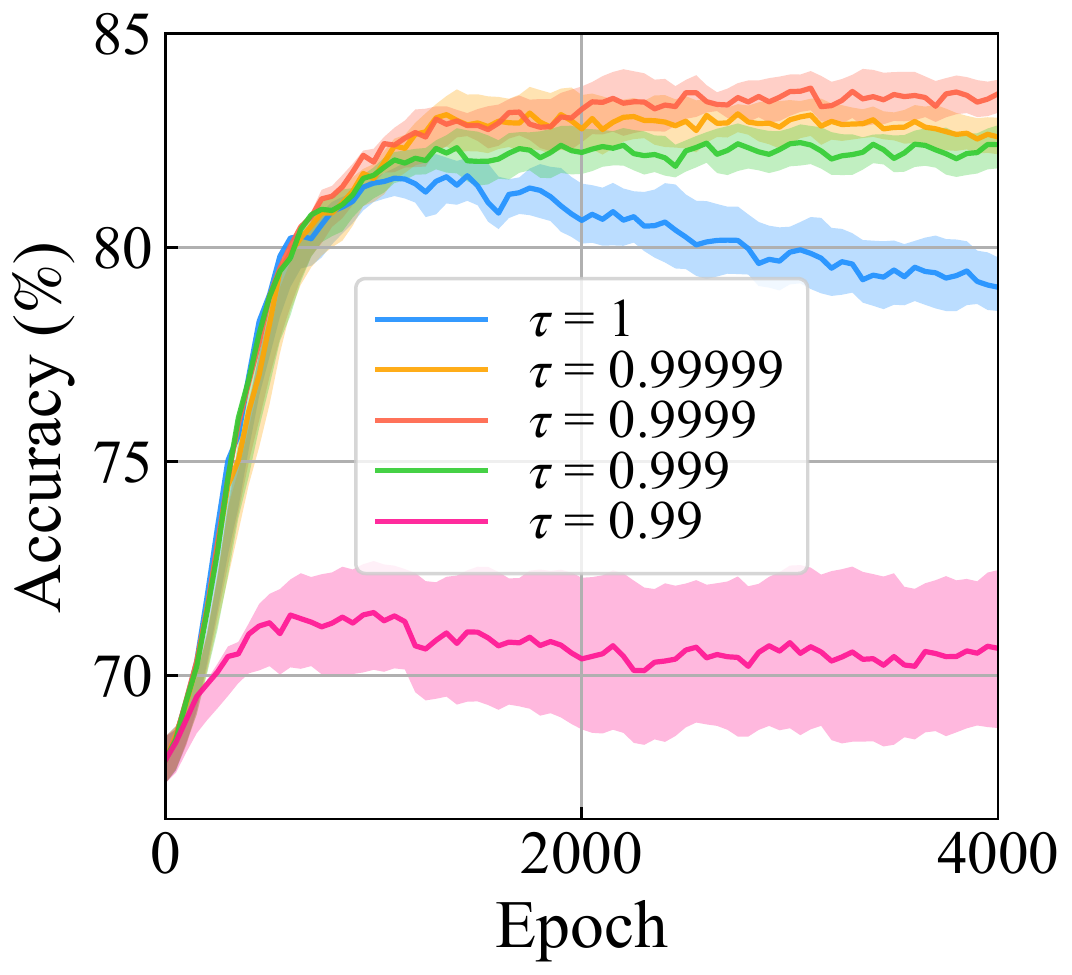}
		\label{fig:acc_curve}
	}\hfill
			\subfigure[Contrastive loss value w.r.t. epoch.]{
		\includegraphics[height=3.5cm]{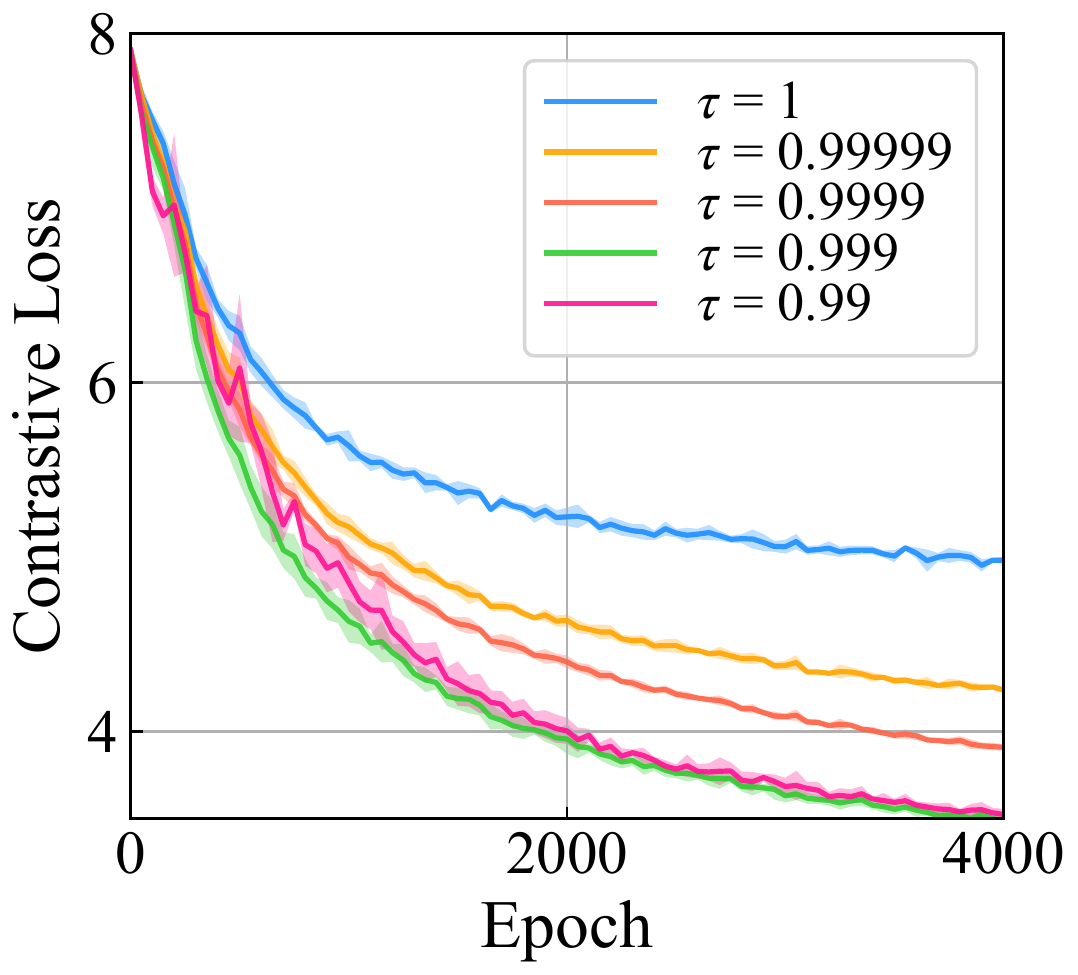}
		\label{fig:loss_curve}
	}
	\vspace{-4mm}
	\caption{Curves of training process on Cora dataset.}
	\label{fig:curves}
    
	\vspace{-4mm}
\end{figure}

In our structure bootstrapping mechanism, the bootstrapping decay rate $\tau$ control the trade-off between updating anchor graph too sharply (with smaller $\tau$) and too slowly (with larger $\tau$). When $\tau = 1$, anchor graph is never updated and remains {as} a constant structure.
To verify the effectiveness of the proposed mechanism, we adjust the value of $\tau$ and the results {are} shown in Table \ref{tab:tau}. We also plot the curves of accuracy and loss value w.r.t. training epoch with different $\tau$, which are shown in Fig. \ref{fig:curves}. 

\begin{table}[t!]
\vspace{-3mm}
\centering
\caption{Test accuracy corresponding to different bootstrapping decay rate $\tau$ in structure refinement scenario.}
\vspace{-4mm}
\begin{tabular}{l|ccccc}
\toprule
\multirow{2}*{\textbf{Dataset}} & \multicolumn{5}{c} {\textbf{Bootstrapping decay rate $\tau$}} \\
	\cline{2-6}
	& ${1}$ & ${0.99999}$  & ${0.9999}$ & ${0.999}$ & ${0.99}$ \\
	\hline
Cora       & 82.1    & 83.2     & \fstHL{84.2}       & 82.4     & 70.9 \\ 
Citeseer   & 71.9    & 72.6     & \fstHL{73.5}       & 73.4     & 72.6 \\ 
Pubmed	   & 80.1    & 80.3     & \fstHL{81.0}       & 80.8     & 80.5 \\ 
\bottomrule
\end{tabular}
\label{tab:tau}
\vspace{-2mm}
\end{table}

As shown in Table \ref{tab:tau}, without structure bootstrapping mechanism ($\tau = 1$), the classification accuracy decreases by $1.5\%$ {on average}, indicating the mechanism helps improve the quality of learned graphs. From Fig. \ref{fig:acc_curve}, we further find an obvious drop after around $1500$ iterations when $\tau = 1$, demonstrating the lack of effective guidance hurts the performance. When $\tau$ is within $[0.999, 0.99999]$, the accuracy can converge to a high value (as shown in Fig. \ref{fig:acc_curve}), meaning that \ourmethod can learn a stable and informative structure with the bootstrapping mechanism. 
However, the performance declines with $\tau$ becoming smaller, especially on Cora dataset. We conjecture that {with sharp updating, the} anchor graph {tends to be polluted} by the {learned} graph {obtained in the early training stage, which fails to} capture {accurate} connections. Another problem caused by a too small $\tau$ is the unstable training, which can be seen in Fig. \ref{fig:acc_curve} and \ref{fig:loss_curve}.

\vspace{-1mm}

\subsection{Sensitivity Analysis (RQ3)}

{Using the} structure inference {case}, we investigate the sensitivity of critical hyper-parameters in \ourmethod, including the probabilities $p^{(x)}$, $p^{(a)}$ for data augmentation and the number of neighbors $k$ in kNN for sparsification and learner initialization. The discussion for $p^{(x)}$ and $k$ are provided below while the analysis for $p^{(a)}$ is given in Appendix \ref{app:sen}. 

\begin{figure}[t!]
    \vspace{-2mm}
	\centering
		\subfigure[Accuracy w.r.t. feature masking rates.]{%
		\includegraphics[height=3.5cm]{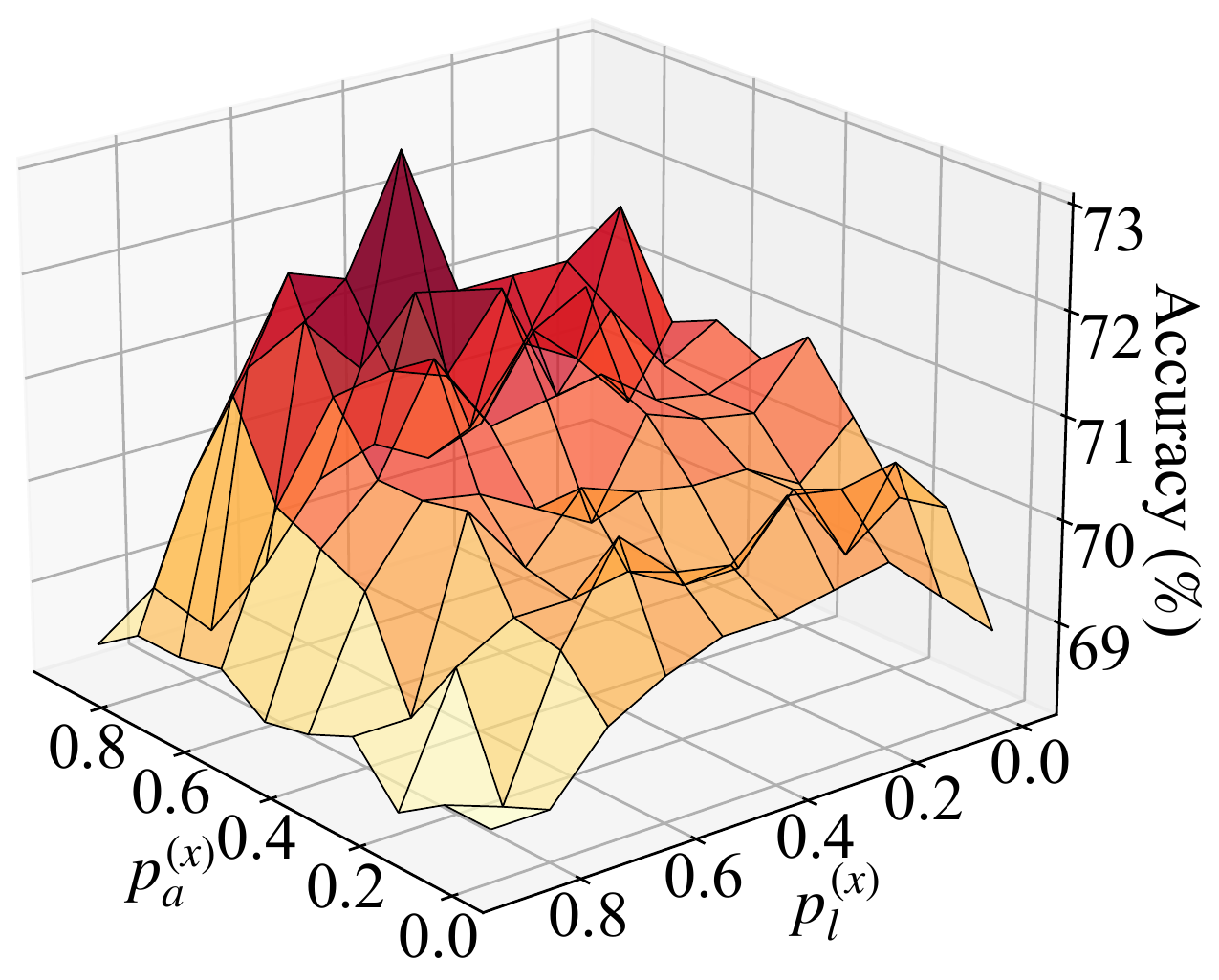}
		\label{fig:param_px}
	}\hfill
		\subfigure[Accuracy w.r.t. number of neighbors.]{
		\includegraphics[height=3.5cm]{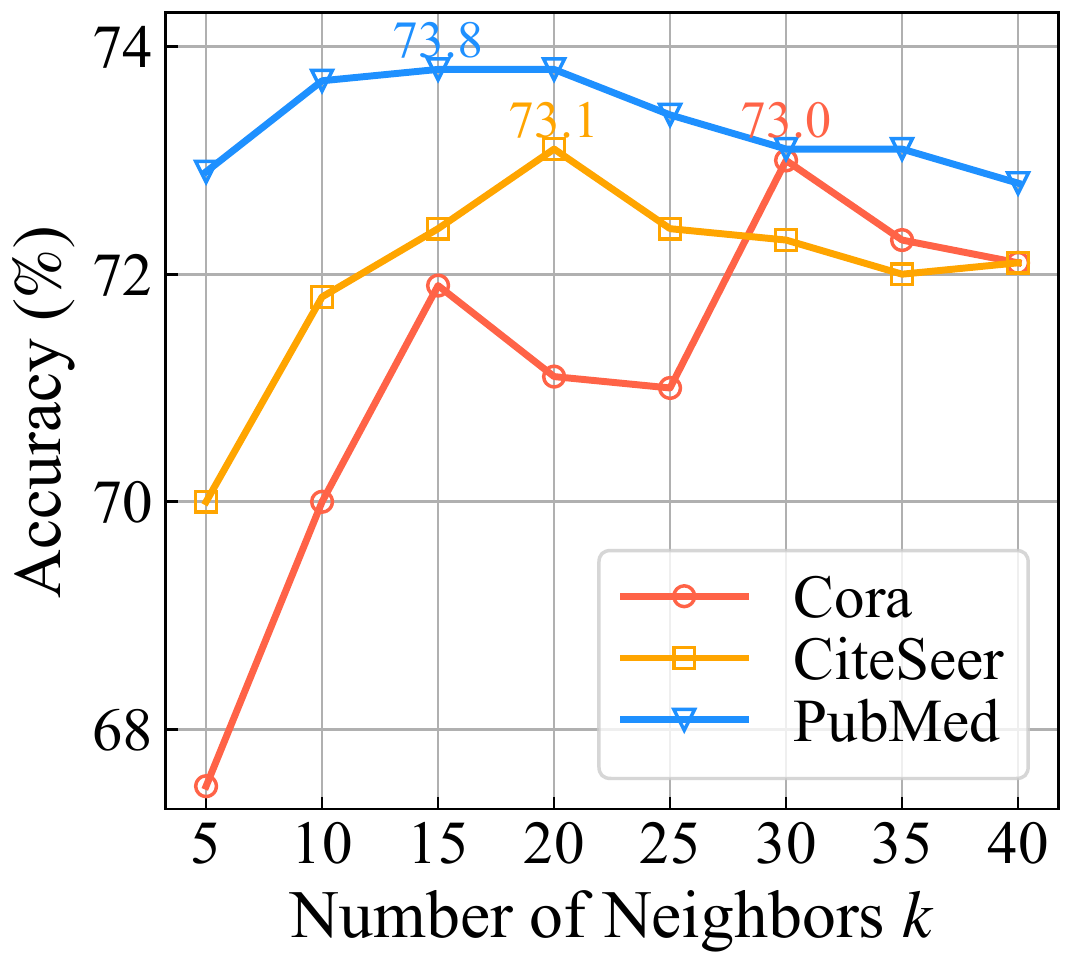}
		\label{fig:param_k}
	}
	\vspace{-4mm}
	\caption{Sensitivity of hyper-parameters $p^{(x)}$ and $k$.}
	\vspace{-4mm}
\end{figure}

\noindent \textbf{Feature masking probability $p^{(x)}$.} 
Fig. \ref{fig:param_px} shows the performance under different combinations of masking probabilities of two views on Cora dataset. We observe that the value of $p^{(x)}_a$ between $0.6$ and $0.8$ produces higher accuracy. Compared to $p^{(x)}_a$, \ourmethod is less sensitive to the choice of $p^{(x)}_l$, suggesting a good performance when $p^{(x)}_l \in [0,0.7]$. When $p^{(x)}$ is larger than $0.8$, the features will be heavily undermined, resulting worse results.

\noindent \textbf{Number of neighbors $k$.} To investigate its sensitivity, we search the number of neighbors $k$ in the range of $\{5, 10, \cdots, 40\}$ for three datasets. As is demonstrated in Fig. \ref{fig:param_k}, the best selection for each dataset is different, i.e., $k=30$ for Cora, $k=20$ for Citeseer, and $k=15$ for Pubmed. A common phenomenon is that a too large or too small $k$ results in poor performance. We conjecture that an extremely small $k$ may limit the number of beneficial neighbors, while an overlarge $k$ causes some noisy connections.

\vspace{-2mm}

\subsection{Robustness Analysis (RQ4)}

\begin{figure}[t!]
\vspace{-2mm}
	\centering
		\subfigure[Accuracy w.r.t. edge deletion rate.]{
		\includegraphics[height=3.5cm]{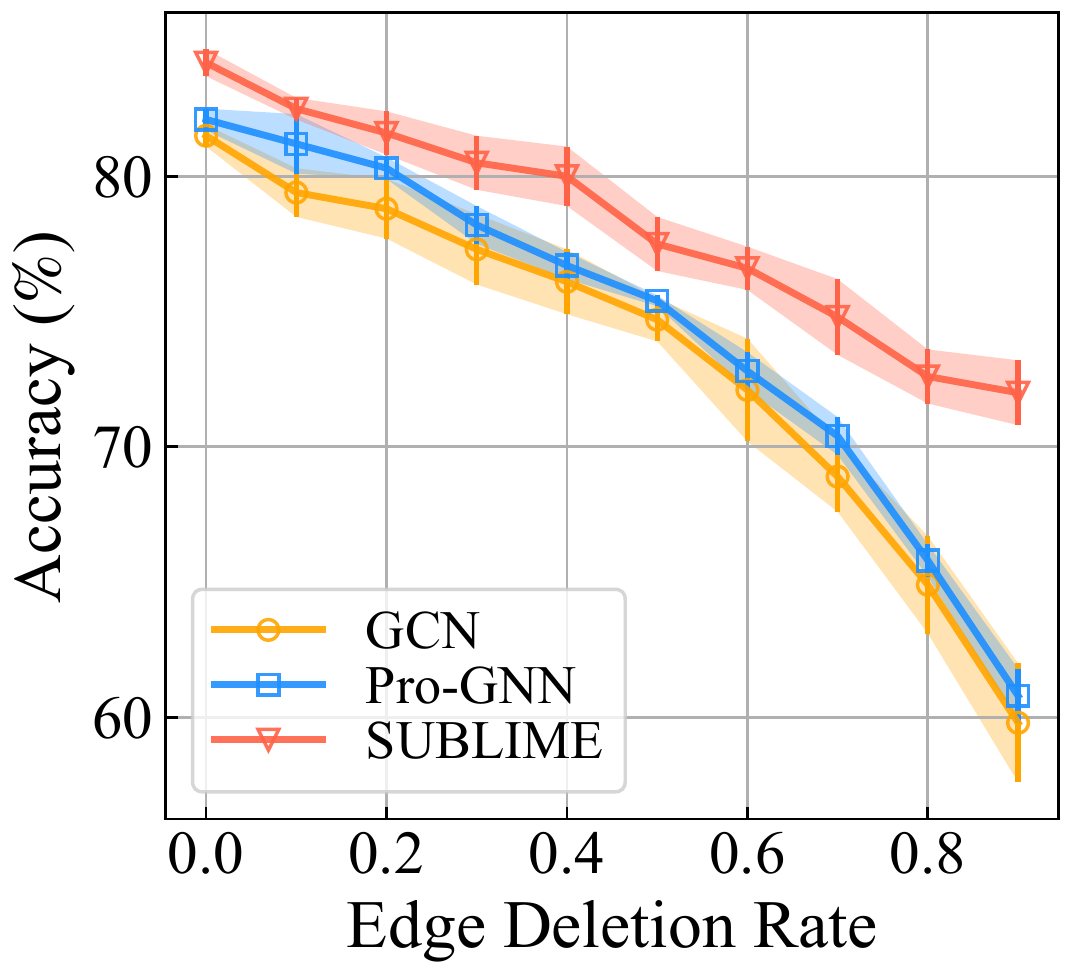}
		\label{fig:dele_edge}
	}\hfill
			\subfigure[Accuracy w.r.t. edge addition rate.]{
		\includegraphics[height=3.5cm]{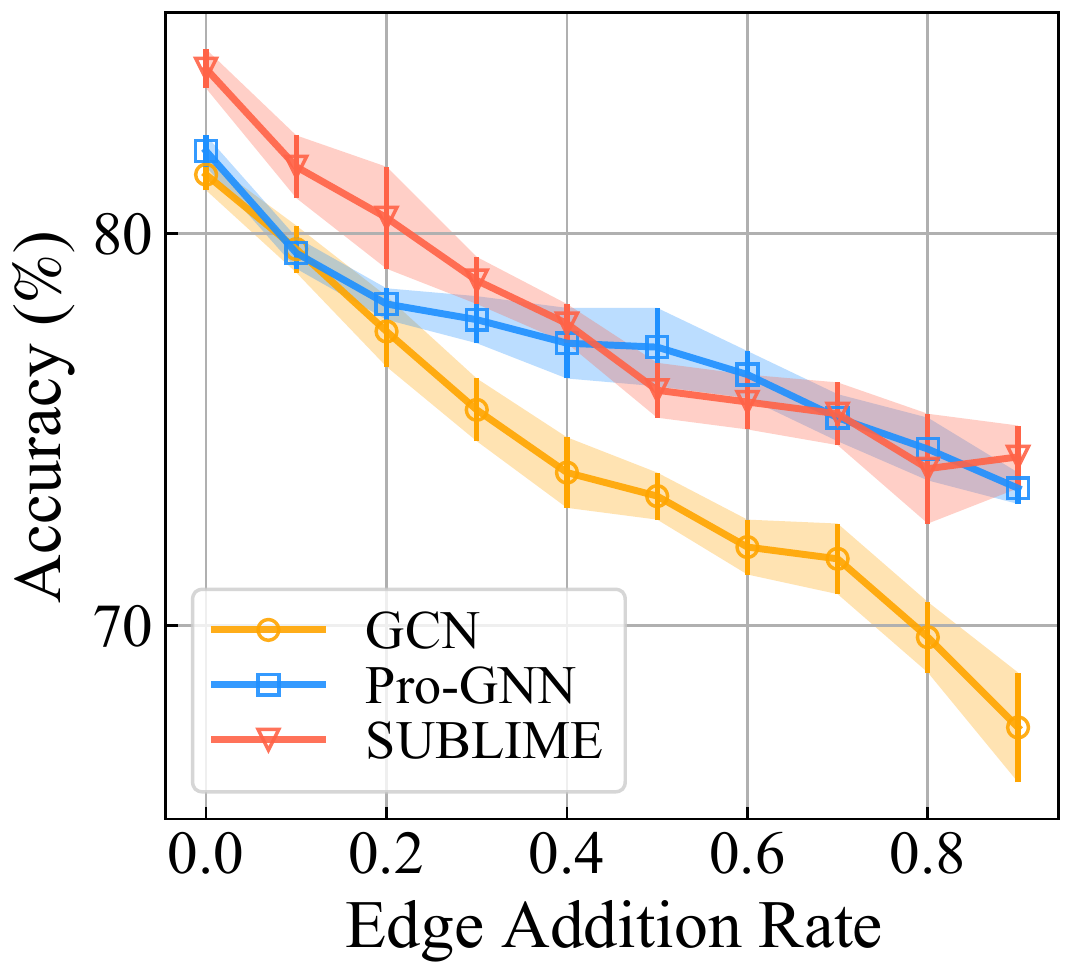}
		\label{fig:add_edge}
	}
	\vspace{-5mm}
	\caption{Test accuracy in the scenarios where graph structures are perturbed by edge deletion or addition.}
	\label{fig:robust}
	\vspace{-1mm}
\end{figure}

To evaluate the robustness of \ourmethod against adversarial graphs, we randomly remove edges from or add edges to the original graph structure of Cora dataset and validate the performance on the corrupted graphs. We change the ratios of modified edges from $0$ to $0.9$ to simulate different attack intensities. We compared our method to GCN \cite{kipf2017semi} and Pro-GNN \cite{jin2020graph}, a supervised graph structure method for graph adversarial defense. As we can see in Fig. \ref{fig:robust}, \ourmethod consistently achieves better or comparable results in both settings. When the edge deletion rates become larger, our method shows more significant performance gains, indicating that \ourmethod has {stronger} robustness against serious structural attacks.

\vspace{-3mm}

\subsection{Visualization (RQ5)}

\newcommand{\mygraphics}{\includegraphics[height=3cm]}
\begin{figure}[t!]
\vspace{-1mm}
\begin{adjustbox}{width=1.\columnwidth,center}
	\centering
		\subfigure[Original graph.]{
		\mygraphics{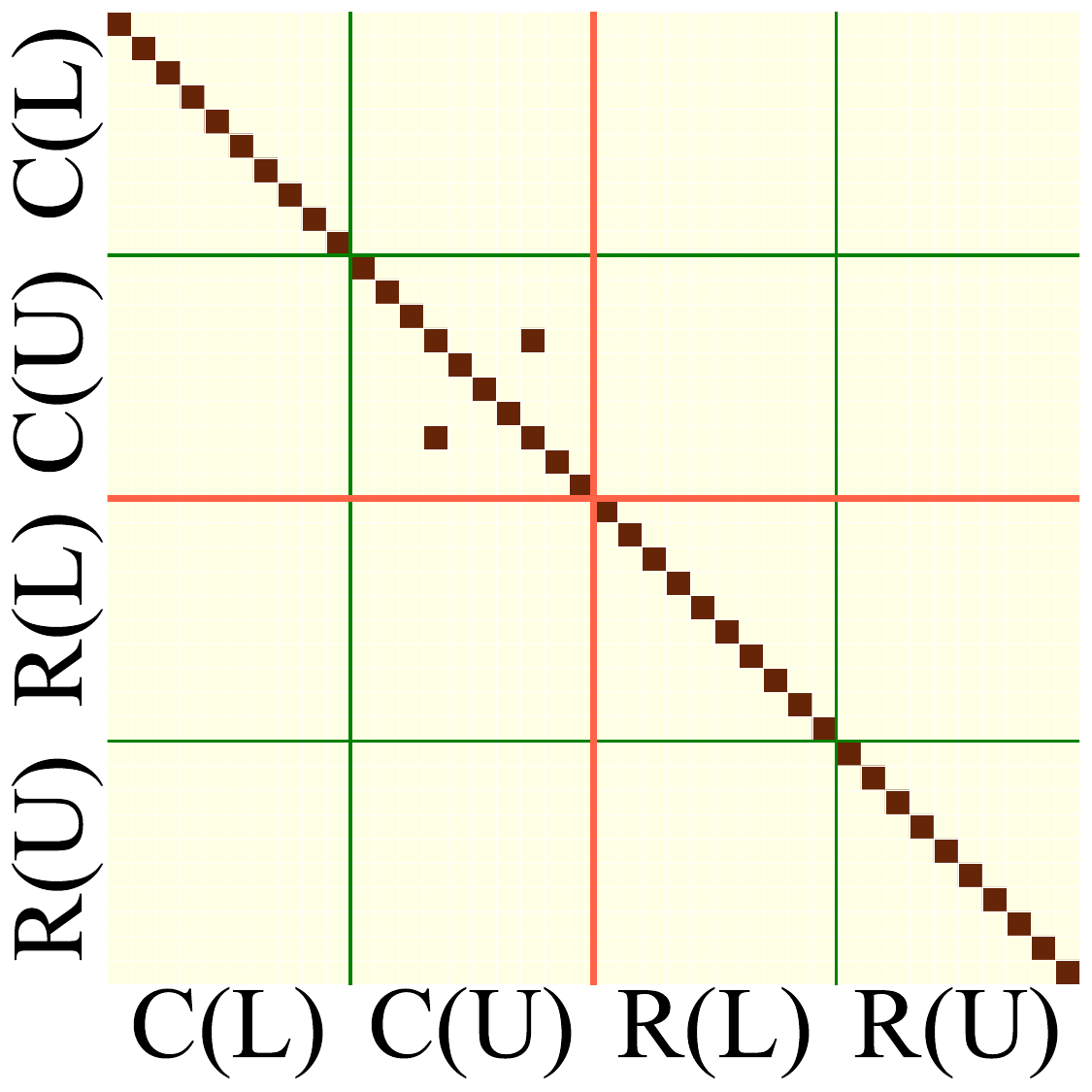}
		\label{fig:hm_origin}
	}\hfill
			\subfigure[Graph learned by Pro-GNN.]{
		\mygraphics{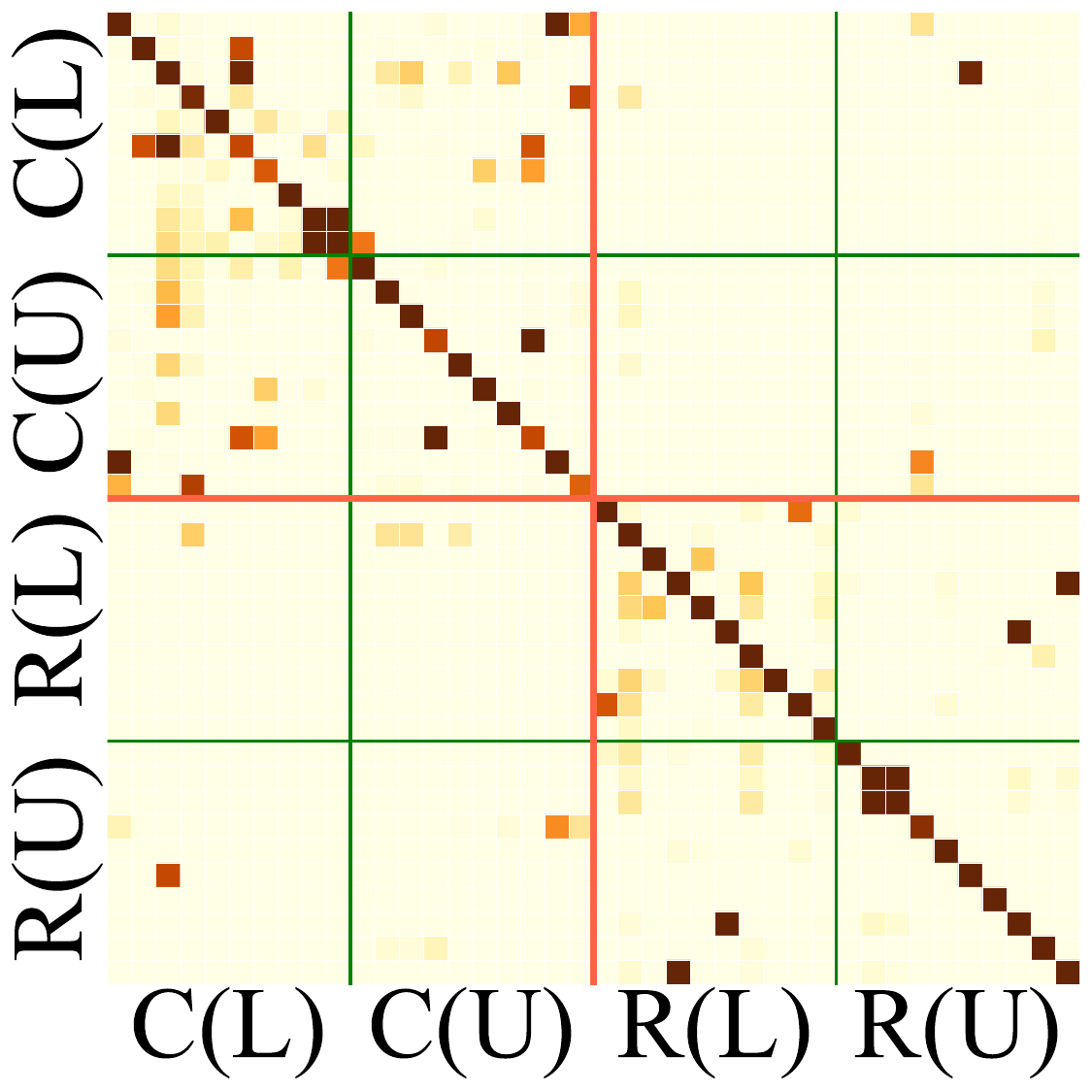}
		\label{fig:hm_learn_pro}
	}\hfill
			\subfigure[Graph learned by \ourmethod.]{
		\mygraphics{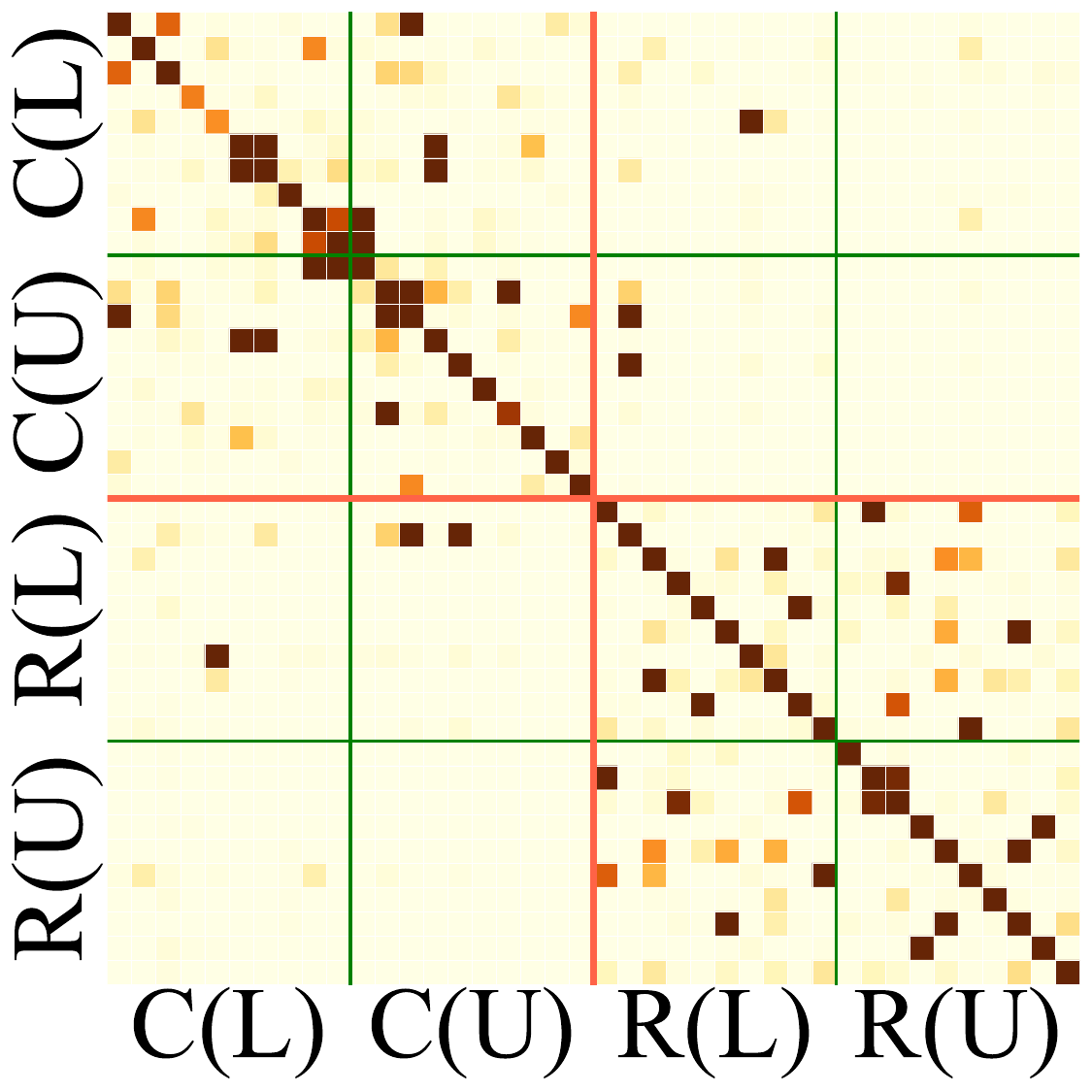}
		\label{fig:hm_learn}
	}
\end{adjustbox}
	\vspace{-5mm}
	\caption{Heatmaps of the subgraph adjacency matrices of (a) the original graph with self-loop, the graph learned by (b) Pro-GNN and (c) \ourmethod on Cora dataset. A block in darker color indicates a larger edge weight between two nodes.}
	\label{fig:heatmap}
	\vspace{-3mm}
\end{figure}

To investigate what kind of graph structure is learned by \ourmethod, we select a subgraph from Cora dataset with nodes in two categories and visualize the edge weights in original graph, graphs {learned} by Pro-GNN and \ourmethod, respectively. The selected categories are Case base (C) and Rule learning (R), each of which has $10$ labeled nodes (L) and $10$ unlabeled nodes (U). Note that the labels of the labeled nodes are used to refine the graph structures in Pro-GNN, but are not used to optimize topology in \ourmethod. As we can see in Fig. \ref{fig:heatmap}, numerous intra-class edges are learned by \ourmethod, while the learned inter-class edges are far fewer than intra-class edges. In contrast, the original graph only provides scarce intra-class edges. We conclude that \ourmethod can learn connections between two nodes sharing similar semantic information, which improves the quality of graph topology. 
Moreover, in Pro-GNN, there are more connections built across labeled nodes than unlabeled nodes, indicating an edge distribution bias in the graph learned by such a supervised method. 
Conversely, \ourmethod equally constructs edges across {all} nodes belonging to the same class as each node can receive the essential supervision from the contrastive {objective}.

\section{Conclusion}
\label{sec:conclusion}
In this paper, we make the first investigation on the problem of unsupervised graph structure learning. To tackle this problem, we design a novel method, \ourmethod, which is capable of leveraging data itself to generate optimal graph structures. To learn graph structures, our method uses contrastive learning to maximize the agreement between the learned topology and a self-enhanced learning target. Extensive experiments demonstrate the superiority of \ourmethod and rationality of the learned structures.

\begin{acks}
The corresponding author is Shirui Pan. This work was supported by an ARC Future Fellowship (No. FT210100097).
\end{acks}

\bibliographystyle{ACM-Reference-Format}
\bibliography{7_references}

\clearpage
\appendix

\section{Notations} \label{ap:notations}

In this paper, we denote scalars with letters (e.g., $k$), column vectors with boldface lowercase letters (e.g., $\mathbf{x}$), matrices with boldface uppercase letters (e.g., $\mathbf{X}$), and sets with calligraphic fonts (e.g., $\mathcal{V}$). The frequently used notations are listed in Table \ref{tab:notations}.

\begin{table}[th]
\vspace{-3mm}
\centering
\caption{Frequently used notations.}
\vspace{-3mm}
\begin{adjustbox}{width=0.99\columnwidth,center}
\begin{tabular}{l|l}
\toprule
\textbf{Notation} & \textbf{Description} \\ \hline
$\mathcal{G} = (\mathbf{A},\mathbf{X})$  &  The (original) graph.\\ 
$n,m,d$  &  The number of nodes/edges/features.\\ 
$\mathbf{A} \in [0, 1]^{n \times n}$ &  The (original) adjacency matrix.\\ 
$\mathbf{X} \in \mathbb{R}^{n \times d}$ &  The feature matrix.\\ 
$\mathcal{G}_l = (\mathbf{S},\mathbf{X})$  &  The learned graph / Learner graph view.\\ 
$\mathbf{S} \in [0, 1]^{n \times n}$  &  The learned adjacency matrix.\\ 
$\tilde{\mathbf{S}} \in \mathbb{R}^{n \times n}$  &  The sketched adjacency matrix.\\ 
$\mathbf{E} \in \mathbb{R}^{n \times d}$  &  The embedding matrix.\\ 
$\mathcal{G}_a = (\mathbf{A}_a,\mathbf{X})$  &  Anchor graph view.\\
$\mathbf{A}_a \in [0, 1]^{n \times n}$ &  The anchor adjacency matrix.\\ 
$\overline{\mathcal{G}}_l, \overline{\mathcal{G}}_a$ & The augmented learner/anchor view. \\
$d_1,d_2$ & The dimension of node representation/projection. \\
$\mathbf{H}_l,\mathbf{H}_a \in \mathbb{R}^{n \times d_1}$ & The representation matrix of learner/anchor view. \\
$\mathbf{Z}_l,\mathbf{Z}_a \in \mathbb{R}^{n \times d_2}$ & The {projected representation} matrix of learner/anchor view. \\
$\mathcal{L}$ & The contrastive loss function. \\ \hline
$p_\omega(\cdot)$ & The graph learner with parameter $\omega$. \\
$q(\cdot)$ & The post-processor. \\
$\mathcal{T}_{fm}(\cdot),\mathcal{T}_{ed}(\cdot)$ & The feature masking/edge dropping augmentation. \\
$f_\theta(\cdot)$ & The GNN-based encoder with parameter $\theta$. \\
$g_\varphi(\cdot)$ & The MLP-based projector with parameter $\varphi$. \\ \hline
$k$ & The number of neighbors in kNN. \\
$p^{(x)}$, $p^{(a)}$ & The masking/dropping probability for $\mathcal{T}_{fm}(\cdot)$/$\mathcal{T}_{ed}(\cdot)$. \\
$\tau, c$ & The decay rate/interval for bootstrapping updating. \\ \hline
$\odot$ & The Hadamard operation. \\
$\cdot^\intercal$ & The transposition operation. \\
\bottomrule
\end{tabular}
\end{adjustbox}
\vspace{-5mm}
\label{tab:notations}
\end{table}

\section{Analysis of Graph Learners} \label{ap:learners}

In Table \ref{tab:learners}, We summarize the properties of the proposed graph learners, including their memory, parameter and time complexity. For metric learning-based graph learners, we consider the complexities with locality-sensitive kNN sparsification post-processing \cite{fatemi2021slaps} where the neighbors are selected from a batch of nodes (batch size $=b_1$). We provide our analysis as follows:

\begin{itemize}
    \item Since FGP learner can model each edge independently and directly, it enjoys several advantages such as the flexibility to model connections and low time complexity. However, its $\mathcal{O}(n^2)$ space complexity makes it hard to be applied to the modeling of large-scale graphs.
    \item Among all metric learning-based learners, attentive learner has the lowest parameter and time complexity w.r.t. dimension $d$. It is suitable for the situation with high feature dimension and low correlation between features.
    \item Compared to attentive learner, MLP and GNN learner require larger space and time complexity to consider the correlation between features and original topology. 
    \item With the effective kNN sparsification, the memory and time complexity are reduced from $\mathcal{O}(n^2)$ to $\mathcal{O}(n)$, which {improves} the scalability of the metric learning-based learners.
\end{itemize}

Considering these properties, we allocate the suitable learner for each dataset. Specifically, for small datasets whose node numbers are less than $3000$ (e.g., Cora), we use FGP learners to model them due to the flexibility and acceptable complexity. For larger datasets with high-dimensional raw features (e.g., Citeseer), we choose attentive learner considering its low parameter/time complexity w.r.t. dimension $d$. For large-scale datasets with relevantly low feature dimensions (e.g., 20news), we adopt MLP learners to capture the correlation between features. In graph refinement scenarios where original graphs are available, GNN learners can be further considered to leverage the extra topology information.

\begin{table}[t!]
\caption{Properties of graph learners.} 
\vspace{-3mm}
\label{tab:learners}
\begin{adjustbox}{width=1.0\columnwidth,center}
\begin{tabular}{lcccc}
\toprule
\textbf{Learner} & \textbf{Sketch} & \textbf{\tabincell{c}{Memory\\ }} & \textbf{\tabincell{c}{Params\\ }} & \textbf{\tabincell{c}{Time\\ }}\\ \midrule
FGP & \begin{minipage}{0.17\textwidth}
      \includegraphics[width=30mm]{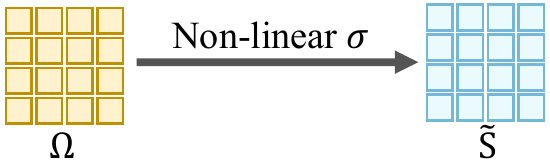}
    \end{minipage} & $\mathcal{O}(n^2)$ & $\mathcal{O}(n^2)$ & $\mathcal{O}(1)$  \\ \midrule
\tabincell{l}{Attentive} & \begin{minipage}{0.17\textwidth}
      \includegraphics[width=30mm]{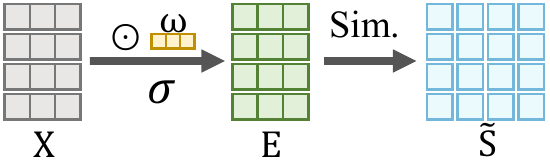}
    \end{minipage} & $\mathcal{O}(ndL +nk)$ & $\mathcal{O}(d L)$ & $\mathcal{O}(ndL+ndb_1)$ \\ \midrule
MLP & \begin{minipage}{0.17\textwidth}
      \includegraphics[width=30mm]{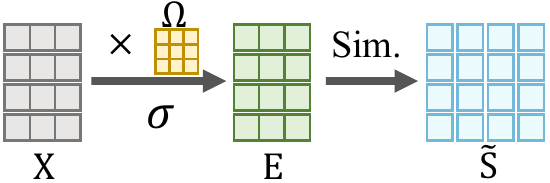}
    \end{minipage} & $\mathcal{O}(ndL +nk)$ & $\mathcal{O}(d^2 L)$ & $\mathcal{O}(nd^2L+ndb_1)$ \\  \midrule
GNN & \begin{minipage}{0.17\textwidth}
      \includegraphics[width=30mm]{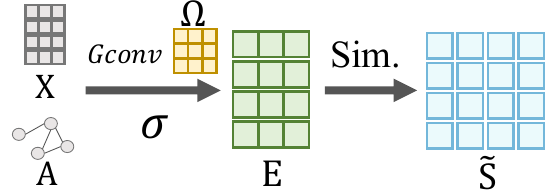}
    \end{minipage} & $\mathcal{O}(ndL +nk)$ & $\mathcal{O}(d^2 L)$ & \tabincell{c}{$\mathcal{O}(mdL+$\\$nd^2L+ndb_1)$ } \\ 

\bottomrule
\end{tabular}
\vspace{-4mm}
\end{adjustbox}
\end{table}

\begin{algorithm}[t]
\caption{The training algorithm of \ourmethod}
\label{alg:overall}
\LinesNumbered
\KwIn{Feature matrix $\mathbf{X}$; Adjacency matrix $\mathbf{A}$ (optional); Number of nearest neighbors $k$; Bootstrapping decay rate and interval $\tau$,$c$; Feature masking probability $p^{(x)}_l$, $p^{(x)}_a$; Edge dropping probability $p^{(a)}$; Temperature $t$; Number of epochs $E$.} 
\KwOut{Learned Adjacency Matrix $\mathbf{S}$}
Initialize parameters $\omega$, $\theta$, $\varphi$;\\
\eIf{$\mathbf{A}$ is provided}{Initialize the anchor adjacency matrix by: $\mathbf{A_a} \leftarrow \mathbf{A}$;}{Initialize the anchor adjacency matrix by: $\mathbf{A_a} \leftarrow \mathbf{I}$;}
\For{$e=1,2,\cdots,E$}{
    Calculate $\tilde{\mathbf{S}}$ with graph learner $p_\omega$ by Eq. (\ref{eq:fgp_learner}) or (\ref{eq:metric_learner});\\
    Calculate ${\mathbf{S}}$ with post-processor $q(\tilde{\mathbf{S}})$ by Eq. (\ref{eq:sparsification}) - Eq. (\ref{eq:normalization});\\
    Establish two graph views by $\mathcal{G}_l = (\mathbf{S},\mathbf{X})$, $\mathcal{G}_a = (\mathbf{A}_a,\mathbf{X})$;\\
    Obtain augmented graph views $\overline{\mathcal{G}}_l$, $\overline{\mathcal{G}}_a$ by Eq. (\ref{eq:maskfeat}) - Eq. (\ref{eq:augmentation}) with probability $p^{(x)}_l$, $p^{(x)}_a$,$p^{(a)}$;\\
    Calculate node representations $\mathbf{H}_l$, $\mathbf{H}_a$ with encoder $f_\theta$ by Eq. (\ref{eq:encoding});\\
    Calculate projections $\mathbf{Z}_l$, $\mathbf{Z}_a$ with encoder $g_\phi$ by Eq. (\ref{eq:projection});\\
    Calculate the contrastive loss $\mathcal{L}$ by Eq. (\ref{eq:cl_loss}) ;\\
    Update parameters $\omega$, $\theta$, $\varphi$ by applying gradient descent; \\
    \If{e mod c = 0}{
    Bootstrapping update $\mathbf{A_a}$ with decay $\tau$ by Eq. (\ref{eq:momentum_update}) ;\\
    }
}
\end{algorithm}

\section{Complexity Analysis} \label{ap:complexity}

We analyze the time complexity of each component of \ourmethod. For graph learner, the complexity has been described in Table \ref{tab:learners}. The time complexity of post-processor is mainly contributed by sparsification, which is $\mathcal{O}(ndb_1)$ for effective kNN and $\mathcal{O}(n^2d)$ for conventional kNN. In the contrastive learning module, the complexities of feature masking and edge dropping are $\mathcal{O}(d)$ and $\mathcal{O}(m)$, respectively. For the encoder and projector, the total complexity is $\mathcal{O}(md_1L_1 + nd_1^2L_1 + nd_2^2L_2)$. For contrastive loss computation, the complexity is $\mathcal{O}(n^2)$ for its full-graph version, $\mathcal{O}(nb_2)$ for the mini-batch version, where $b_2$ is the batch size of contrastive learning. 

\section{Algorithm} \label{ap:algo}

The training algorithm of \ourmethod is summarized in Algorithm \ref{alg:overall}. 

\begin{table}[t!]
\centering
\caption{Statistics of datasets.}
\vspace{-3mm}
\begin{adjustbox}{width=0.95\columnwidth,center}
\begin{tabular}{l|ccccc}
\toprule
\textbf{Dataset} & \textbf{Nodes} & \textbf{Edges}  & \textbf{Classes} & \textbf{Features} & \textbf{\tabincell{c}{Label Rate}} \\ \hline
Cora       & 2,708    & 5,429     & 7       & 1,433     & 0.052 \\ 
Citeseer   & 3,327    & 4,732     & 6       & 3,703     & 0.036 \\ 
Pubmed	   & 19,717   & 44,338    & 3       & 500       & 0.003 \\ 
ogbn-arxiv & 169,343  & 1,166,243 & 40      & 128       & 0.537 \\ 
Wine       & 178      & N/A       & 3       & 13        & 0.056 \\ 
Cancer     & 569      & N/A       & 2       & 30        & 0.018 \\ 
Digits     & 1,797    & N/A       & 10      & 64        & 0.028 \\ 
20news     & 9,607    & N/A       & 10      & 236       & 0.010 \\ 

\bottomrule
\end{tabular}
\end{adjustbox}
\label{tab:dataset}
\vspace{-4mm}
\end{table}

\section{Datasets} \label{app:dataset}

In Table \ref{tab:dataset}, we summarize the statistics of benchmark datasets. The dataset splitting follows the previous works \cite{franceschi2019learning,chen2020iterative}. Details of these datasets are introduced as follows.

\begin{itemize}
    \item \textbf{Cora} \cite{sen2008collective} is a citation network where each node is a machine learning paper belonging to $7$ research topics and each edge is a citation between papers. 
    \item \textbf{Citeseer} \cite{sen2008collective} is a citation network containing $6$ types of machine learning papers: Agents, AI, DB, IR, ML, and HCI. Nodes denote papers and edges denote citation relationships.
    \item \textbf{Pubmed} \cite{pubmed_namata2012query} is a citation network from the PubMed database, where nodes are papers about three diabete types about diabetes and edges are citations among them. 
    \item \textbf{ogbn-arxiv} \cite{hu2020ogb} is a citation network with Computer Science arXiv papers. 
    The features are the embeddings of words in its title and abstract. The labels are 40 subject areas.
    \item \textbf{Wine} \cite{asuncion2007uci} is a non-graph dataset containing the results of a chemical analysis of $178$ wines derived from three different cultivars. Features are the quantities of 13 constituents found.
    \item \textbf{Cancer} \cite{asuncion2007uci} is a binary classification dataset of diagnosis of breast tissues (malignant/benign). The features are computed from a digitized image of a breast mass.
    \item \textbf{Digits} \cite{asuncion2007uci} is a non-graph dataset containing handwritten digits in 10 classes. Each sample is a $8\times8$ image of a digit.
    \item \textbf{20news} \cite{asuncion2007uci} is a non-graph dataset comprising newsgroups posts on $20$ topics. Following \cite{franceschi2019learning}, we select $10$ topics with $9,607$ samples in our experiments. 
\end{itemize}

\section{Implementation Details} \label{app:exp_detail}

\subsection{Computing Infrastructures} 

We implement \ourmethod using PyTorch 1.7.1 \cite{paszke2019pytorch} and DGL 0.7.1 \cite{wang2019dgl}. All experiments are conducted on a Linux server with an Intel Xeon 4214R CPU and four Quadro RTX 6000 GPUs.

\subsection{Evaluation Details}

Through node classification tasks, we evaluate the quality of the learned structures by re-training a classifier with the learned structure as its constant input. Specifically, we use the learned adjacency matrices to train GCN-based classification models, and record the testing result with the highest validation accuracy. The averaged accuracy {over} five {rounds of running} is used to assess the classification {performance}. For ogbn-arxiv dataset, we utilize a three-layer GCN with $256$ hidden units as the evaluation model. For the rest datasets, a two-layer GCN with $32$ hidden units is employed. 

For node clustering tasks, we evaluate the performance of our method by measuring the quality of the learned representations. Concretely, following the baseline methods \cite{agc_zhang2019attributed,daegc_wang2019attributed}, we train our framework for a fixed number of epochs and apply K-means algorithm for $10$ runs to group the learned representations. The representations are generated by the contrastive learning encoder $f_\theta$ taking learned graph $\mathcal{G}_l=(\mathbf{S},\mathbf{X})$ as its input without augmentation.

\begin{figure}[!t]
  \includegraphics[height=4.1cm]{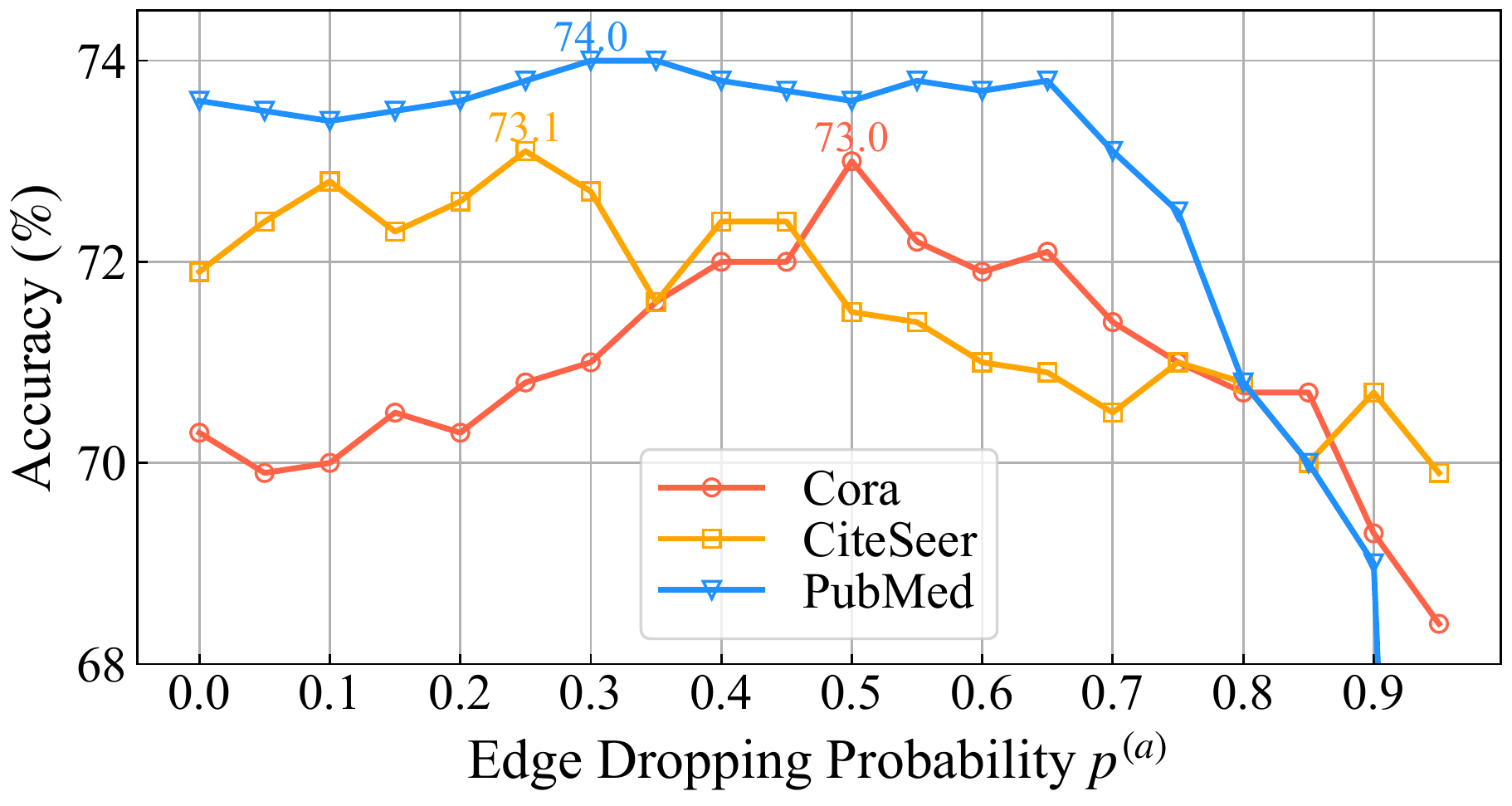}
  \vspace{-4mm}
  \caption{Sensitivity analysis for $p^{(a)}$.}
  \label{fig:param_pa}
  \vspace{-4mm}
\end{figure}

\subsection{Hyper-parameter Specifications}

We perform grid search to select hyper-parameters on the following searching space: the dimension of representation and projection is searched in $\{16, 32, 64, 128, 256, 512\}$; $k$ on kNN is tuned amongst $\{5, 10, 15, 20, 25, 30, 35, 40\}$; feature masking probability $p^{(x)}$ is tuned from $0.1$ to $0.9$; edge dropping probability $p^{(a)}$ is searched in $\{0,0.25,0.5,0.75\}$; the bootstrapping decay rate is chosen from $\{0.99,0.999,0.9999,0.99999,1\}$; and the learning rate of Adam optimizer is selected from $\{0.01, 0.001, 0.0001\}$. The temperature for contrastive loss is fixed to $0.2$. The layer numbers of encoder ($L_1$), projector ($L_2$), and embedding network ($L$) are set to $2$.

For our baselines, we reproduce the experiments using their official open-source codes or borrow the reported results in their papers. We carefully tune their hyper-parameters to achieve optimal performance. To compare fairly, we use the random seeds $\{0,1,2,3,4\}$ for all classification methods, and fix the seed to $0$ for clustering methods.

\section{Parameter Sensitivity of $p^{(a)}$} \label{app:sen}

We vary the dropping rate $p^{(a)}$ from $0$ to $0.95$ on Cora, Citeseer and Pubmed datasets, and the results are shown in Fig. \ref{fig:param_pa}. As we can see, when $p^{(a)}$ is between $0.2$ and $0.65$, \ourmethod achieves better performance. When the edge dropping rate is overlarge, the structures on both views {will be deteriorated}, causing a sharp drop of performance.

\end{document}